\documentclass[twoside]{article}

\usepackage[accepted]{aistats2024}

\usepackage{amsmath,amsfonts,bm}

\def\eqref#1{equation~\ref{#1}}

\def\1{\bm{1}}

\def\ra{{\textnormal{a}}}

\def\rs{{\textnormal{s}}}

\def\rx{{\textnormal{x}}}
\def\ry{{\textnormal{y}}}

\def\rvw{{\mathbf{w}}}
\def\rvx{{\mathbf{x}}}
\def\rvy{{\mathbf{y}}}
\def\rvz{{\mathbf{z}}}

\def\rmH{{\mathbf{H}}}

\def\vmu{{\bm{\mu}}}
\def\vtheta{{\bm{\theta}}}
\def\vphi{{\bm{\phi}}}

\def\mC{{\bm{C}}}

\def\mH{{\bm{H}}}

\DeclareMathAlphabet{\mathsfit}{\encodingdefault}{\sfdefault}{m}{sl}
\SetMathAlphabet{\mathsfit}{bold}{\encodingdefault}{\sfdefault}{bx}{n}

\newcommand{\E}{\mathbb{E}}

\newcommand{\Ccal}{\mathcal{C}}
\newcommand{\Dcal}{\mathcal{D}}

\newcommand{\Fcal}{\mathcal{F}}

\newcommand{\Ncal}{\mathcal{N}}

\newcommand{\Xcal}{\mathcal{X}}

\newcommand{\EE}{\mathbb{E}} %

\newcommand{\masa}[1]{\noindent{\textcolor{purple}{\{{\bf Masa:} \em #1\}}}}

\DeclareMathOperator*{\argmax}{arg\,max}
\DeclareMathOperator*{\argmin}{arg\,min}
\newcommand{\newedit}{\color{blue}}

\usepackage{natbib}
\usepackage{url}
\usepackage{subcaption}
\usepackage{algorithm}
\usepackage{algorithmic}
\usepackage{amssymb}
\usepackage[dvipsnames]{xcolor}
\usepackage{enumerate}
\usepackage{comment}
\usepackage{amsthm}
\usepackage{apxproof}
\usepackage{thmtools, thm-restate}
\usepackage{wrapfig}
\usepackage{adjustbox}
\usepackage[flushleft]{threeparttable}
\usepackage{graphicx}
\usepackage{tcolorbox}
\usepackage[page, header]{appendix}
\usepackage{titletoc}
\usepackage{lipsum}
\usepackage{hyperref}
\usepackage{multirow}
\usepackage{booktabs}
\usepackage{threeparttable}
\usepackage{enumitem}
\usepackage{mathabx}
\usepackage{subcaption}
\hypersetup{
    citecolor=Blue,
    colorlinks=true,
    linkcolor=WildStrawberry,
    filecolor=magenta,      
    urlcolor=blue,
linktocpage}
\usepackage{comment}

\DeclareMathAlphabet\mathbfcal{OMS}{cmsy}{b}{n}

\newcommand{\mytag}[2]{%
  \text{#1}%
  \@bsphack
  \protected@write\@auxout{}%
         {\string\newlabel{#2}{{#1}{\thepage}}}%
  \@esphack
}
\usepackage{hyperref}

\begin{document}

\twocolumn[

\aistatstitle{Functional Graphical Models: \\ Structure Enables  Offline Data-Driven Optimization}

\aistatsauthor{ Jakub Grudzien Kuba \And Masatoshi Uehara  \And  Pieter Abbeel \And Sergey Levine }

\aistatsaddress{ BAIR, UC Berkeley \And Genentech \And BAIR, UC Berkeley  \And BAIR, UC Berkeley } ]

\begin{abstract}
{
\vspace{-20pt}
While machine learning models are typically trained to solve prediction problems, we might often want to use them for optimization problems. For example, given a dataset of proteins and their corresponding fluorescence levels, we might want to optimize for a new protein with the highest possible fluorescence. This kind of data-driven optimization (DDO) presents a range of challenges beyond those in standard prediction problems, since we need models that successfully predict the performance of new designs that are better than the best designs seen in the training set.
It is not clear theoretically when existing approaches can even perform better than the na\"{i}ve approach that simply selects the best design in the dataset. In this paper, we study how structure can enable sample-efficient data-driven optimization. To formalize the notion of structure, we introduce functional graphical models (FGMs) and show theoretically how they can provide for principled data-driven optimization by decomposing the original high-dimensional optimization problem into smaller sub-problems. This allows us to derive much more practical regret bounds for DDO, and the result implies that DDO with FGMs can achieve nearly optimal designs 
in situations where na\"ive approaches fail due to insufficient coverage of the offline data. We further present a data-driven optimization algorithm that inferes the FGM structure itself, either over the original input variables or a latent variable representation of the inputs.}
\end{abstract}

\section{INTRODUCTION}
\vspace{-10pt}

Machine learning models are typically trained for prediction, but we often want to use such models to solve \emph{optimization} or \emph{decision-making} problems. Imagine we would like to use a dataset of fluorescent proteins and their fluorescence levels to design a new protein with the highest possible fluorescence~\citep{brookes2019conditioning, trabucco2022design}. Similarly, when provided with examples of hardware accelerators and their performance, an engineer might want to infer a more performant hardware accelerator design~\citep{kao2020confuciux, kumar2021data}. 
A direct machine learning approach to such problems would be to train a surrogate model that predicts the performance for a given design, and then find its maximizers with some optimization method~\citep{trabucco2021conservative}. In this work, we focus on the challenging yet practical scenario where we can use offline training data to train a surrogate model, but cannot collect additional samples online.
This offline data-driven optimization (DDO) approach is often applied in model-based optimization \citep[MBO]{hoburg2014geometric, gomez2018automatic, trabucco2021conservative} and offline reinforcement learning \citep[RL]{levine2020offline, fujimoto2021minimalist}. 


The primary challenge in DDO is 
insufficient data coverage of optimal examples, which causes the distribution shift in the optimization process: when we optimize the design (\textit{i.e.}, the input to the model), we produce a design very different from the training distribution. Predictions for such designs are likely to be inaccurate, and an optimizer might exploit the model to generate designs that result in the largest (optimistic) errors. Consequently, recent works in MBO and offline RL have focused on incorporating penalties for exploiting data outside of offline data when constructing a surrogate model \citep{kumar2020conservative, trabucco2021conservative}. However, even with these conservative approaches, it remains unclear whether we can learn the optimal design when the offline training data fails to cover it. The question in this context is: \textit{What enables DDO to be feasible in this challenging offline scenario?}

{We make the observation that sample-efficient DDO can be enabled by leveraging structure. Without any structural bias for the surrogate model, even with the aforementioned conservative methods, a na\"ive approach in the worst case reduces to choosing the best design observed in the offline data, as we will formalize in Section~\ref{sec:pre}. As the design space becomes large, the probability of an optimal design being actually present in the offline data generally becomes extremely low.}

{ In our work, we introduce the framework of \emph{Functional Graphical Models (FGMs)} as a structure that facilitates sample-efficient DDO. FGMs are designed to describe function's independence properties, allowing us to decompose every function into subfunctions over partially intersecting cliques. Importantly, FGMs provide an effective way to introduce structural bias for DDO by automatically decomposing the whole optimization problem into smaller subproblems. We demonstrate that DDO with FGMs can mitigate distribution shift, as it only requires the optimal design to be covered by the offline data in a more lenient manner, compared to DDO without any structural bias. Specifically, FGMs enable the learning of the optimal design as long as the variables in each \textit{clique} of the FGM take on optimal values for some data point---it is not required to see any single point with optimal values for \emph{all} cliques jointly. As a result, we can formally demonstrate that DDO with FGMs surpasses the na\"ive approach. 
}



Our contributions can be summarized as follows. First, we introduce FGMs and show how they give rise to a natural decomposition of the function of interest
over itsx cliques. 
Second, we demonstrate that FGMs can reduce regret
in data-driven optimization significantly. In particular, our results imply that we can learn a near-optimal
design as long as it is covered by the offline data within individual cliques of the FGM, rather than the entire space. Lastly, we propose a practical data-driven optimization algorithm that can discover FGMs from the offline data, under Gaussian assumptions, or from a learned latent space in a more general setting. We validate the effectiveness of our algorithm in combination with MBO through numerical experiments that demonstrate the benefit of the discovered FGMs, especially for high-dimensional functions with many cliques.




\vspace{-5pt}
\section{RELATED WORK}
\vspace{-5pt}

Data-driven optimization has connections to several fields, including Bayesian optimization and model-based optimization, as summarized below. 

\vspace{-5pt}
\paragraph{Bayesian Optimization.} The idea of using data and machine learning tools for design and optimization has been one of the primary motivations of \textit{Bayesian optimization} (BO) \citep{brochu2010tutorial, shahriari2015taking}. 
In BO, one establishes a \textit{prior} belief of the considered function and updates it based on the given dataset, as well as queries for function evaluations that balance maximization and exploration of the function. 
However, such an \textit{online} approach is not applicable to problems where additional queries cannot be made or are very cost- and time-consuming (\textit{e.g.}, drug design) that would ideally be solved fully \textit{offline}.
Nevertheless, we consider our work to be of interest to the BO community, where it is common to \textit{assume} a decomposition of the objective over a \textit{dependency graph}, which simplifies the problem and enables solving it with fewer queries \citep{kandasamy2015high, rolland2018high, hoang2018decentralized, ziomek2023random}. 
In Section \ref{sec:sgs}, we existence of such a decomposition for every function under mild conditions, enabling tractable optimization with finite amounts of offline data. 

\vspace{-5pt}
\paragraph{Probabilistic Graphical Models.} 
Learning FGMs involves uncovering the structure of functions that map inputs $x$ to outputs $y$. Similarly, the challenge of discovering structural graphs has been extensively studied in the field of probabilistic graphical models \citep[PGMs]{jordan1999learning, studeny2006probabilistic, pearl2009causality}. FGMs and PGMs differ because PGMs are built upon probabilistic independence, while FGMs are constructed based on the concept of \emph{functional independence},
which we will introduce later in Section \ref{sec:sgs}. In the literature on PGMs, the main focus is on uncovering statistical relationships between variables, whereas for the DDO problems that we study, the main goal of using FGMs is to enable efficient optimization with fewer samples, as we will formalize in Section~\ref{sec:regret}. 


\vspace{-5pt}
\paragraph{Model-Based Optimization.}

{
There has been a growing interest in model-based optimization (MBO) and the specific instantiation of MBO from static data, or data-driven optimization (DDO). Recent works have investigated computational design via such \textit{offline} MBO for chemical molecules, proteins, hardware accelerators, and more \citep{gomez2018automatic, brookes2019conditioning, kumar2020model, kumar2021data}. MBO and DDO are often used interchangeably, and we employ the term DDO to refer specifically to optimization from static \emph{offline} data. Most of these prior methods adopt a ``conservative'' (\emph{i.e.}, pessimistic) approach to DDO, employing regularizers to prevent the generation of out-of-distribution designs \citep{trabucco2021conservative, qi2022data}. 
Combined with appropriate inductive bias, these methods found application in practical engineering problems, such as hardware accelerator (microchip) design \citep{kumar2021data}.  
In our paper, we instead investigate how discovering functional independence structure via our FGM framework can enable efficient optimization from data, and provide to our knowledge the first theoretical analysis of sample efficiency for DDO with such independence properties.
}

\vspace{-5pt}
\section{PRELIMINARIES}\label{sec:pre}
\vspace{-5pt}
In DDO, the algorithm receives an offline dataset $\mathcal{D}=\{ \rvx^{(i)}, \ry^{(i)}\}_{i=1}^{N}$, where the input $\rvx \in \mathcal{X}$ is drawn from a data distribution $p\in \Delta(\Xcal)$, and $\ry=f(\rvx) \in\mathbb{R}$ is the corresponding evaluation of an unknown (black-box) function $f$ at $\rvx$. Our goal is to find an input (\textit{i.e.}, design) $\rvx$ that takes a high value in terms of $f(\rvx)$. In the ideal case, we may want to find an exact $\rvx^{\star}\triangleq\argmax_{\rvx \in \Xcal}f(\rvx)$. 
In our analysis, we study a generalization of this problem, where DDO learns a stochastic distribution $\pi \in \Delta(\Xcal)$ that maximizes $J(\pi)$, where $J(\pi)$ is the value defined by $J(\pi)=\E_{\rvx\sim \pi}[f(\rvx)]$. 


{
In DDO (\textit{e.g.}, MBO), conventional methods typically involve the following steps: (a) obtaining a surrogate model $\hat f:\Xcal \to \mathbb{R}$ using function approximation techniques like deep neural networks, and (b) optimizing the surrogate model with respect to $\rvx \in \Xcal$. This is formalized in the following procedure:
\begin{align} \label{eq:naive2}
\hat f =&\argmin_{\bar f \in \mathcal{F}}\E_{(\rvx,\ry)\sim \mathcal{D}}[\{\bar f(\rvx)-\ry\}^2] , \tag{Step (a)}\\ 
& \quad \hat \pi = \argmax_{\pi \in \Pi}\E_{\rvx \sim \pi}[\hat f(\rvx)], \tag{Step (b)} \label{eq:naive}
\end{align}
where $\Pi$ consists of probability distributions over $\Xcal$, and $\Fcal$ is a regression class that consists of functions mapping $\Xcal$ to $\mathbb{R}$. When $\Pi$ is a fully expressive class\footnote{When $\Xcal$ is discrete, we can always use this fully expressive class. However, when $\Xcal$ is continuous, one must use a restricted distribution class, \textit{e.g.,} $\Pi=\{\Ncal(\rvx,\sigma^2);\rvx\in \Xcal$\} for small $\sigma^2$.}, i.e., $\Pi=\Delta(\Xcal)$, Step (b) reduces to 
$$\textstyle
\hat \rvx = \argmax_{\rvx \in \Xcal} \hat f(\rvx).
$$
The primary practical challenge in the fully offline DDO setting lies in dealing with insufficient data coverage. To see this, consider the na\"{i}ve approach: we can set a model $\hat f(\rvx)$ to match the observed $f(\rvx)$ values and assign random values to examples outside of the offline data. This approach amounts to constructing a surrogate model $\hat f$ in the initial Step (a) without relying on any structural bias. 
In the discrete case, it is equivalent to setting 
$\Fcal=\{\rvx \mapsto \vtheta^{\top}\vphi(\rvx) \ | \ \vtheta \in \mathbb{R}^{|\mathcal{X}|}\}$, where $\phi(\rvx)$ is a $|\mathcal{X}|$-dimensional one-hot encoding vector. 
Maximizing such a model corresponds to choosing $\rvx$ for which $\theta_{\rvx}$ is highest. However, if the input space $\Xcal$ is not fully covered by the data, such an optimization process may end up outputting a novel design $\rvx$ whose ground-truth value $f(\rvx)$ is very low, but its random prediction $\hat{f}(\rvx)$ happens to be high, thus adversarially exploiting the model's errors.
One may want to tackle this problem by, instead of assigning random values to unobserved designs, modeling them \textit{pessimistically}---for example, by setting them to $\hat{f}(\rvx)=-\infty$.
The resulting solution is: 
\begin{align}\label{eq:simple}
   \hat \rvx =\argmax_{\rvx \in \Xcal}\hat f(\rvx)=\argmax_{\rvx \in \Dcal}\hat f(\rvx),
\end{align}
which implies that the performance of the output design is bounded by the performance of the examples in the offline data.
Hence, in either case, the performance of DDO is expected to be poor when the data distribution $p(\rvx)$ does not cover regions with high output values in terms of $f(\rvx)$. This issue is particularly severe when the space $\Xcal$ is large, which tends to cause $p(\rvx^{\star})$ to be low. 

To tackle this problem 
our work explores reasonable structural bias that can be incorporated into surrogate model classes $\Fcal$, mitigates the problem of insufficient coverage, and can be learned from the offline data.
}

\paragraph{Notation.} We write $\mathcal{V}$ to denote the index set of input variables.
If $\mathcal{X}\subseteq\mathbb{R}^{d}$, then $\mathcal{V}=[d]\triangleq \{1, \dots, d\}$. Thus, $\rvx_{\mathcal{V}}$ and $\rvx$ are equivalent. For a subset $\mathcal{S}\subseteq \mathcal{V}$, we denote $\rvx_{\mathcal{S}}$ as the vector of variables with indexes in $\mathcal{S}$ and $\Xcal_{\mathcal{S}}$ as its domain. We denote the rest of the vector in $\rvx$ by $\rvx_{-\mathcal{S}}$. Furthermore, $a \lesssim  b$ stands for inequality up to a problem-dependent constant. The notation $\Ncal(\mu,\sigma^2)$ denotes a normal distribution with mean $\mu$ and variance $\sigma^2$, and we use $\Ncal$ to denote $\Ncal(0, I)$ for simplicity.


\vspace{-5pt}
\section{FUNCTIONAL GRAPHICAL MODELS}
\vspace{-5pt}
\label{sec:sgs}

We present the concept of \emph{functional independence}, which characterizes the structural relationship between inputs and outputs. We then build on this definition to introduce a graphical representation of functional independences between variables, which we refer to  as \emph{Functional Graphical Models (FGMs)}. Then, we will show how to apply these concepts to DDO. 


\vspace{-5pt}
\subsection{Functional Independence}
\vspace{-5pt}

We define the concept of functional independence, which decomposes the structure of functions mapping from $\Xcal$ to real numbers. This concept can be used as a useful inductive bias when constructing surrogate functions for DDO.

\begin{restatable}[Functional Independence]{definition}{difind}
    \label{def:difind}
    Let $A, B, S \subseteq \mathcal{V}$.  We say that $\rvx_{A}$ and $\rvx_{B}$ are functionally independent given $\rvx_{S}$ if, there exist functions $f_{-(B\setminus S)}$ and $f_{-(A\setminus S)}$, such that for every $\rvx\in\mathcal{X}$,
        \begin{align}
            f(\rvx) = f_{-(B\setminus S)}(\rvx_{\mathcal{V}\setminus (B\setminus S)}) + f_{-(A\setminus S)}(\rvx_{\mathcal{V}\setminus (A\setminus S)}). \nonumber
        \end{align}
    If $S=\emptyset$ and there exist mutually disjoint supersets $\underline{A}$ of $A$ and $\underline{B}$ of $B$ such that $f_{-B}(\rvx_{-B}) = f_{\underline{A}}(\rvx_{\underline{A}})$, and $f_{-A}(\rvx_{-A}) = f_{\underline{B}}(\rvx_{\underline{B}})$ the independence is absolute. 
\end{restatable}

Intuitively, functional independence of $\rvx_A$ and $\rvx_B$ given $\rvx_S$ states that, once we fix the value of $\rvx_S$, the two sets of variables can be optimized independently. This is a desirable property for DDO because 
each input subspace is covered denser by the dataset if viewed individually rather than jointly with another subspace.
Intuitively, this allows an optimization algorithm to decompose the original problem into two smaller ones with more effective data. 
We will formalize this intuition in Section~\ref{sec:regret}.

Our first result offers alternative criteria for defining functional independence when $\Xcal$ is contained in an Euclidian space. These properties play a key role in discovering structure in unknown functions, as we will discuss in Section~\ref{sec:discovery}. 

\begin{restatable}{lemma}{difindcond}
    \label{lemma:criteria}
    Suppose $f(\rvx)$ is twice-continuously differentiable w.r.t. $\rvx$. 
    Let $A, B, S\subseteq \mathcal{V}$. Then, the following statements are equivalent.
    \begin{enumerate}
        \item $\rvx_A$ and $\rvx_B$ are functionally independent given $\rvx_S$. 
        \item For every $\rvx\in\mathcal{X}$, 
    \begin{align}\frac{\partial f}{\partial \rvx_{A\setminus S}} (\rvx) = F_{-(B\setminus S)}(\rvx_{\mathcal{V}\setminus (B \setminus S)}),\nonumber
    \end{align}
    for some function $F_{-(B\setminus S)}(\rvx_{\mathcal{V}\setminus (B \setminus S)})$ that does not take $\rvx_{(B\setminus S)}$ as input.
        \item For every $\rvx\in\mathcal{X}$,
        \begin{align}
            \frac{\partial^2 f}{\partial \rvx_{A\setminus S} \partial \rvx_{B\setminus S}} (\rvx) = 0.\nonumber
        \end{align}
    \end{enumerate}
\end{restatable} 
Detailed proofs are provided in Appendix~\ref{ape:proof}.

In the following subsection, we introduce a graphical representation of functional independence that enables encoding structures within high-dimensional functions and leads to a provably efficient approach to DDO.
\begin{figure}[!t]
  \vspace{-5pt}
  \centering
    \includegraphics[width=0.5\linewidth]{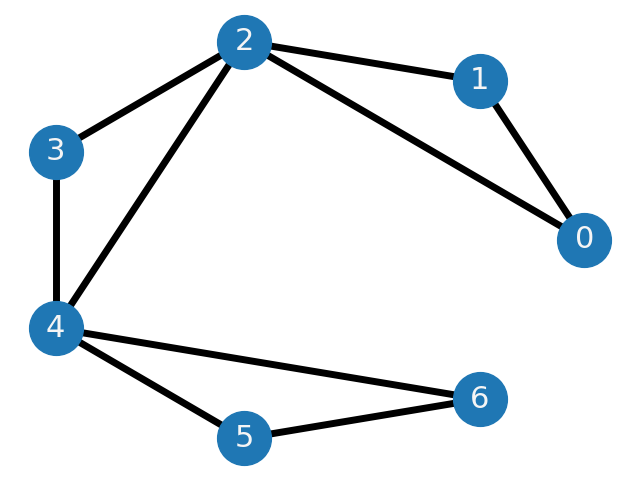}
  \caption{ {\small The clique set of this graph is given by $\{ \{0, 1, 2\}, \{2, 3, 4\}, \{4, 5, 6\} \}$. Hence, if it is a FGM of $f(\rvx)$, then $f(\rvx)=f_{0,1,2}(\rvx_{0,1,2}) + f_{2,3,4}(\rvx_{2,3,4}) + f_{4,5,6}(\rvx_{4,5,6})$.}}
  \vspace{-5pt}
  \label{fig:example}
\end{figure}
\vspace{-5pt}
\subsection{FGMs}
\vspace{-5pt}

We first define functional graphical models (FGMs), a general framework for working with functional independence. An example is shown in Figure~\ref{fig:example}.
\begin{restatable}[Functional Graphical Model (FGM)]{definition}{FGM}
    \label{def:sg}
    A graph $\mathcal{G}=(\mathcal{V}, \mathcal{E})$ is an FGM of a function $f(\rvx)$ if, for any $i, j\in\mathcal{V}$ such that $i\neq j$, $(i,j)\notin \mathcal{E}$, we have that $\rx_i$ and $\rx_j$ are functionally independent given $\rvx_{\mathcal{V}\setminus\{i, j\}}$. FGM with the smallest set $\mathcal{E}$ is referred to as the minimal FGM of $f$.
\end{restatable}

Our construction of FGMs
is analogous to that of probabilistic graphical models (PGMs), where the absence of an edge indicates probabilistic conditional independence between two random variables \citep{jordan1999learning, koller2009probabilistic}. 
As we prove in Appendix \ref{appendix:ours}, a result similar to the Hammersley-Clifford theorem in PGMs \citep{clifford1971markov} holds for FGMs, decomposing arbitrary functions as follows. 

\begin{restatable}[Function Decomposition]{theorem}{decomp}
    \label{theorem:decomp}
    Assumue that $\int_{\mathcal{X}}\exp f(\rvx) d\rvx$ exists and let $\mathcal{G}$ be any FGM of $f(\rvx)$ and $\mathcal{C}$ be its set of maximal cliques (clique set). Then, there exist functions $\{f_{C}(\rvx_{C}), C\in\mathcal{C}\}$, such that
    \begin{align}
        f(\rvx) = \sum_{C\in\mathcal{C}} f_{C}(\rvx_C).\nonumber
    \end{align}
\end{restatable}

We provide an illustrative example of application of this theorem in Figure \ref{fig:example}.

\vspace{-5pt}
\subsection{DDO via FGMs}
\vspace{-5pt}
\label{subsec:MBO_FGM}


{
Consider a DDO algorithm that incorporates FGMs by decomposing its surrogate model, following Theorem \ref{theorem:decomp}, over the maximal cliques of the FGM of $f$.
Then, DDO can be instantiated as follows:
\begin{align*}
 \hat \pi_{\mathrm{FGM}} &= \argmax_{\pi \in \Pi}\E_{\rvx \sim \pi}[\hat f(\rvx)],\,\hat f(\rvx)= \sum_{C\in \mathcal{C}} \hat f_C(\rvx_C), \\
 \{\hat f_C\}_{C\in \mathcal{C}} &=\argmin_{\{\bar f_C \in \mathcal{F}_C\}_{C\in \mathcal{C}} }\E_{(\rvx,\ry)\sim \mathcal{D}}\Big[\Big(\sum_{C\in \mathcal{C}} \bar f_C(\rvx_C)-y\Big)^2\Big],
\end{align*}
where each $\Fcal_C$ is a function class that consists of maps from $\Xcal_C$ to $\mathbb{R}$. 

Now, consider an illustrative simple case where $\Xcal$ is discrete.
In this case, natural choices of $\Pi$ and $\Fcal_C$ would be fully expressive classes: $\Pi=\Delta(\Pi),\Fcal_C=\{\rvx_C\mapsto \vtheta_{C}^{\top} \vphi_C(\rvx_C) \ | \ \vtheta_C \in \mathbb{R}^{|\Xcal_C|}  \}$ where $\vphi_C$ is a one-hot encoding vector over $\Xcal_C$. Then, the abovementioned method is equivalent to choosing 
\begin{align}\label{eq:simple2}
    \argmax_{\overline{\rvx} \in \Xcal}\{ \vphi^{\top}&(\overline{\rvx})\E_{\rvx \sim \Dcal}[ \vphi(\rvx) \vphi(\rvx)^{\top}]^{-1}\E_{(\rvx,\ry)\sim \Dcal}[\vphi(\rvx)\ry]\}, \nonumber\\
    &\vphi(\rvx) = [\vphi(\rvx_1)^{\top},\cdots,\vphi(\rvx_C)^{\top}]^{\top}.
\end{align}
Importantly, this method differs from just choosing an optimal $\rvx$ in the dataset $\Dcal$, i.e., $\hat \rvx_{\mathrm{sim}}$ in Eq. (\ref{eq:simple}) which corresponds to a method without any inductive bias. 

In practice, we must still estimate FGMs from the data and handle the continuous space $\Xcal$. We will delve into the practical implementation of this in Section~\ref{sec:discovery}.
}

\vspace{-5pt}
\section{FGMs ENABLE SAMPLE-EFFICIENT DATA-DRIVEN OPTIMIZATION}
\vspace{-5pt}
\label{sec:regret} 

In this section, we present our main result, demonstrating that functions represented by FGMs can be optimized efficiently with much less stringent data coverage assumptions than those needed for standard data-driven optimization. \footnote{Note our theory in this section still holds when an observation $\ry$ has a bounded measurement error $\epsilon$, i.e., $\ry=f(\rvx) + \epsilon$.} 


\subsection{REGRET GUARANTEES} \label{subsec:regret}
\vspace{-0.2cm}
First, we review the regret guarantee of the na\"ive approach in \ref{eq:naive2} and \ref{eq:naive}. A standard regret guarantee in this setting~\citep{chen2019information} states that, with probability $1-\delta$, the following holds:
\begin{align}\label{eq:regret_naive}
  \underbrace{J(\pi^{\star}) - J(\hat \pi)}_{\mathrm{Regret\,against}\,\pi^{\star} }
  \lesssim \underbrace{\max_{\bar \pi\in \Pi}\max_{\rvx \in \mathcal{X}}\left |\frac{\bar \pi(\rvx)}{p(\rvx)}\right |}_{\mathrm{Coverage (a)}}\times \underbrace{\sqrt{\frac{\log(|\mathcal{F}|/\delta)}{n}}}_{\mathrm{Complexity (a')}}, 
\end{align}
where $\pi^{\star}$ is an optimal design among $\Pi$, i.e., $\pi^{\star}=\argmax_{\pi \in \Pi}J(\pi)$.  

The coverage term (a) measures the discrepancy between a distribution in $\Pi$ and a data distribution $p$.
The term (a') signifies the size of the function class $\mathcal{F}$. Note this $|\mathcal{F}|$ could be easily replaced with the covering number when $\Fcal$ is infinite \citep{wainright2019high}. 

Our main result shows that DDO with FGMs enjoys a much more favorable regret guarantee:
\begin{restatable}[Regret of DDO with FGMs]{theorem}{regretfgm}
\label{thm:regret}
Let $\Ccal$ be the set of maximal cliques of an FGM of $f(\rvx)$ and $f(\rvx)=\sum_{C\in\mathcal{C}}f_{C}(\rvx_C)$ be its FGM decomposition.
Under the following assumptions $\forall C\in\Ccal$,

(1) the function approximator classes are centered    
\begin{align}
    \E_{\rvx\sim p}[\bar f_C(\rvx_C)]=0, \forall \bar f_C \in \mathcal{F}_C,\nonumber
\end{align}
(2) the correlations between cliques are well-controlled
\begin{align}
    \max_{\{\bar f_C\in \mathcal{F}_C\} }\mathrm{Corr}[\bar f_{C_1}(\rvx_{C_1}),\bar f_{C_2}(\rvx_{C_2})]\leq \sigma, \nonumber
\end{align}
(3) the models are well-specified, sp that $f_C \in \mathcal{F}_C$, and 

(4) $|\bar f_C(\rvx_C)|\leq 1$ for any $\rvx_C \in \mathcal{X}_C, \bar f_C\in \Fcal_C$,

the following result holds with probability $1-\delta$
\begin{align}
  &J(\pi^{\star}) - J(\hat \pi_{\mathrm{FGM}})\lesssim \nonumber\\
  &\underbrace{\sqrt{\frac{1}{1-\sigma}}}_{\mathrm{Corr} (b'')}\underbrace{\max_{C\in \mathcal{C}}\max_{\bar \pi\in \Pi}\max_{\rvx_C \in \mathcal{X_C}}\left |\frac{\bar \pi(\rvx_C)}{p(\rvx_C)}\right |}_{\mathrm{Coverage (b)}} \underbrace{\sqrt{\frac{|\mathcal{C}|\sum_{C}\log(|\mathcal{F}_C|/\delta)}{n}}}_{\mathrm{Complexity (b')}}, \nonumber
\end{align}
where $p(\rvx_C),\bar \pi(\rvx_c)$ denote the marginal distributions of  $\rvx_C$ with respect to $p(\rvx)$ and $\pi(\rvx)$.
\end{restatable}

\begin{figure}[!t]
  \centering
\includegraphics[width=0.65
\linewidth]{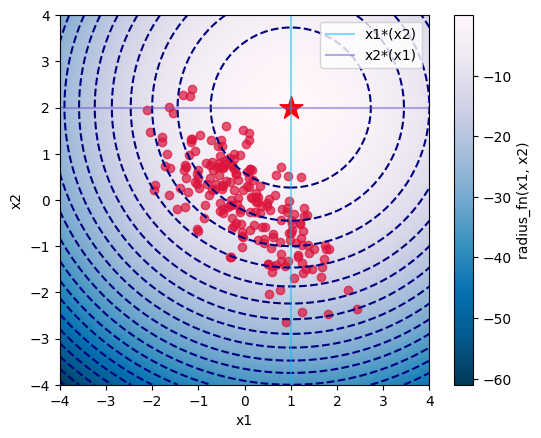}
  \caption{Consider a function $f(\rx_1, \rx_2)=-(\rx_1-1)^2 -(\rx_2-2)^2$. Clearly, the singleton cliques $\{\rx_1\}$ and $\{\rx_2\}$ are functionally independent, while the data coming from a correlated normal distribution are not statistically independent.
  While the dataset does not jointly cover the optimal solution $\rvx^{\star}=(1,2)$, it does cover individual components $\rx^{\star}_{1}=1, \rx^{\star}_{2}=2$, and thus a method that can learn the component functions can compose them into $\rvx^{\star}$.}
  \label{fig:data_dist}
\end{figure}

While the bound of the na\"{i}ve approach in Eq. (\ref{eq:regret_naive}) implies that MBO without FGM structure necessitates the data distribution $p(\rvx)$ to cover the optimal design itself, the FGM factorization only necessitates the data distribution to cover components within each \textit{clique} in the optimal design.
This is a significantly less stringent requirement, as the best design in the offline data might be much less effective than the best design achievable by combining favorable settings of the values in each of the cliques, as illustrated in Figure~\ref{fig:data_dist}, and analyzed deeper below.



\vspace{-0.2cm}
\paragraph{Coverage term.} The term (b) corresponds to coverage that takes into account the FGM factorization. This term quantifies how a distribution $\pi \in \Pi$ differs from a data distribution $p$
in terms of \emph{marginals} with respect to $\rvx_C$. The following lemma demonstrates that term (b) is smaller than the coverage term (a) across the entire space $\Xcal$:

\begin{restatable}[Improvement of Coverage terms]{lemma}{lemmacoverage}
\label{lem:helpful}
For any $\pi\in \Delta(\Xcal)$, we have 
\begin{align}
\max_{C \in \mathcal{C}}\max_{\rvx_C \in \Xcal_C}\frac{\pi(\rvx_C)}{p(\rvx_C)} \leq    \max_{\rvx\in \Xcal }\frac{\pi(\rvx)}{p(\rvx)}. \nonumber
\end{align} 
\end{restatable}

Consider the following illustrative example: suppose the space $\Xcal$ is discrete, denoted by $\Xcal = \Xcal_1 \otimes \Xcal_2 \otimes \cdots\otimes \Xcal_d$ where $\Xcal_1=\Xcal_2=\cdots =\Xcal_d$. Now, let's assume that $f(\rvx) = \sum_C f_C(\rvx_{C})$, and $p(\rx_1, \rx_2, \cdots, \rx_d)$ follows a jointly independent uniform distribution. Term (a) exhibits exponential growth with the dimension $d$: 
\begin{align}
\label{eq:explode}
   \max_{\bar \pi\in \Delta(\Xcal)} \max_{\rvx \in \Xcal }\frac{\bar \pi(\rx_1,\rx_2,\cdots,\rx_d)}{p(\rx_1,\rx_2,\cdots,\rx_d)}=|\Xcal_1|^d. 
\end{align}
Therefore, the na\"{i}ve data-driven optimization procedure suffers from a curse of dimensionality, requiring an exponentially large number of samples to obtain a near-optimal design as the dimension increases. On the contrary, the term (b) remains reasonably small:
\begin{align}\label{eq:not_explode}
   \max_{\bar \pi\in \Delta(\Xcal)}\max_{C \in \mathcal{C}} \max_{\rvx_C\in \Xcal_C } \frac{\bar \pi( \rvx_C)}{p(\rvx_C)}\leq |\Xcal_1|^{\max_{C} d_C} 
\end{align}
where $d_C$ represents the dimension of $\rvx_C$. Thus, the DDO with FGMs can effectively overcome the curse of dimensionality when $\max_{C} d_C$ is reasonably small.

Furthermore, in a more extreme example, illustrated in Figure~\ref{fig:data_dist}, the coverage term (b) is bounded, but the coverage term (a) is infinite. This illustrates that DDO with FGMs can identify the optimal design, when DDO without FGMs cannot do so even with infinite samples. It's worth noting that in this scenario, a discerning reader might express concerns about the potential explosion of the additional term $(b'')$, which could adversely affect the regret of DDO with FGMs. However, as we will discuss in Section~\ref{subsec:notes}, this term $(b'')$ is moderately constrained. 


\vspace{-0.2cm}
\paragraph{Statistical complexity term.} We can anticipate that $(b')$ is generally much smaller than $(a')$
because the function class for the surrogate model has more structure. Let's examine a specific scenario, where $\Xcal_1=\cdots =\Xcal_d$ and each $\Xcal_i$ is discrete. Suppose that $\mathcal{F}$ is a linear combination of all polynomial basis functions over $[\rx_1,\cdots,\rx_d]$, i.e., the fully expressive class. In this case, a standard argument based on covering \citep{wainright2019high} shows that the term $(a')$ would be on the order of $O(\sqrt{|\Xcal_1|^d/n})$, because the number of parameters in the model is $|\Xcal|=|\Xcal_1|^d$. 
Now, let's consider the term $(b')$. When we set each $\mathcal{F}_C$ to be a linear combination of all polynomial basis functions over the entire $\Xcal_C$, its order becomes $O(\sqrt{|\mathcal{C}||\Xcal_1|^{\max_C d_{C}}/n})$, because the number of parameters in the model is bounded from above by $|\mathcal{C}||\Xcal_1|^{\max_C d_{C}}$. Again, this implies that DDO with FGMs can effectively overcome the curse of dimensionality when $\max_{C} d_C$ is reasonably small.

\vspace{-5pt}
\paragraph{Summary.}

Theorem~\ref{thm:regret} reveals that the original data-driven optimization problem naturally breaks down into subproblems for each clique $C \in \mathcal{C}$. More specifically, this result shows that, while the na\"{i}ve approach requires the data distribution to cover the optimal design, with FGMs, it is only necessary for the data distribution to cover the optimal values of each clique separately. With non-trivial FGMs, as we see in Eq. (\ref{eq:explode}) and Eq. (\ref{eq:not_explode}),
the difference in the number of samples needed to obtain a near-optimal design between DDO with FGMs and na\"{i}ve DDO (i.e., DDO without FGMs) can be exponential \footnote{Note the term (b") is still $1$ in this case. Hence, we can ignore the term (b") for the comparison.}.
{
Furthermore, although there could be a potential concern related to the additional term $(b'')$, in certain instances, DDO with FGMs has the capability to learn the optimal design, whereas the na\"{i}ve DDO approach cannot achieve this even with an infinite amount of data.} 
The intuition for this is that na\"{i}ve DDO requires the samples from the data distribution to randomly set \emph{all} of the variables to near-optimal values in at least some of the training points in the offline data, whereas DDO with FGMs only requires some training points for each clique to have near-optimal values, and does not require any single training point to have near-optimal values for \emph{all} cliques. In the na\"{i}ve case, this can lead to catastrophic sample complexity, essentially reducing the method to selecting the best point in the offline data (i.e., the na\"{i}ve approach in Eq.~\ref{eq:simple}), whereas with FGMs and relatively small cliques, it is possible to recombine the best values for each clique and find good designs potentially with exponentially fewer samples.

  
Note that the literature on additive models \citep{hastie1987generalized} has previously explored their role in reducing statistical complexity. In contrast, we characterize the coverage term in the context of DDO. It's important to highlight that simply reducing statistical complexity polynomially (i.e., improving the term $(a')$) is insufficient for overcoming the curse of dimensionality, because the term $(a)$ might still grow exponentially.

\subsection{Assumptions on DDO with FGMs }\label{subsec:notes}



Theorem \ref{thm:regret} shows that DDO with FMGs enjoys favorable regret guarantees under a set of assumptions of the DDO problem. In this section, we’ll discuss these assumptions in more detail to provide the reader with intuition about our main result
The first and last assumptions are primarily technical. 
The first one can always be satisfied with negligible error by substituting $f_C(\rvx^{(i)})$ and $\ry^{(i)}$ with $f_C(\rvx^{(i)})-\E_{\rvx_{C}\sim\Dcal}[f(\rvx_C)]$ and $\ry^{(i)}-\E_{\rvy \sim \Dcal}[\ry]$. 
The third assumption is standard, and we can easily account for potential misspecification errors in the theorem. The second assumption is substantial. When $\sigma=1$, our guarantee is essentially void. In essence, this assumption implies that each clique exhibits mild independence, automatically breaking down the entire problem into subproblems. Technically, this enables us to translate the error of $\hat f(\rvx)=\sum_C \hat f_C(\rvx_C)$ for the entire graph into the error of $\hat f_C$ for each individual clique. 

Readers might have concerns about whether the second assumption could potentially lead to issues, particularly in cases where cliques overlap. 
In this context, it's important to note that no two cliques overlap completely, nor is one entirely contained within another, since they are elements of the set of maximal cliques of the FGM. 
Furthermore, this assumption can be significantly relaxed. In the second assumption, while the maximum is taken over the entire function class $\mathcal{F}_C$, we can replace it with a much smaller set consistent with $f(\rvx)$, as we demonstrate in Appendix \ref{appendix:experiments}. Secondly, in many cases, we can still observe huge improvements in terms of regret, especially when the FGM cliques are much smaller than the whole graph. We illustrate this empirically in Section \ref{sec:experiments}.

\vspace{-5pt} 
\subsection{Relation with Pessimism}
\vspace{-5pt} \label{sec:pessmism}

Most of the common recent approaches to offline DDO can be characterized by \textit{pessimism}, displayed in algorithms through, for example, additional losses that penalize exploring out-of-distribution designs \citep{wu2019behavior, kumar2020conservative, trabucco2021conservative}.
To facilitate the comparison between the na\"{i}ve procedure without FGMs and the one with FGMs, in our work, we did not include pessimistic penalties for searching outside of offline data in the algorithms. When we incorporate pessimism into the algorithms, we can formally show that it alleviates distribution shift in that we can obtain a regret bound where $\max_{\bar \pi \in \Pi}$ in the coverage term is replaced with just an optimal design $\pi^{\star}$ \citep{rashidinejad2021bridging,xie2021bellman}. In this sense, pessimism alleviates the distributional shift problem. However, in the instance we show, even if we replace $\max_{\bar \pi\in \Pi}$ with an optimal design $\pi^{\star}$,
the coverage term in Eq. (\ref{eq:explode}) for the procedure without
FGMs is still $O(|\Xcal_1|^d)$, and the coverage term in Eq. (\ref{eq:not_explode}) for the procedure with FGMs is still $O(|\Xcal_1|^{\max_C d_C})$. Consequently, pessimism itself cannot resolve distribution shift when the optimal design is not well-covered by the data distribution $p(\rvx)$. On the contrary, FGMs relax the requirement for the data coverage of the whole design space $\Xcal$ into a much milder one of coverage of subspaces induced by individual cliques.

\vspace{-5pt}
\subsection{Relation with Offline RL} 
\vspace{-5pt}
The decomposition provided by FGMs inspires thoughts about their connection to offline RL.
As we alluded to before, offline RL can be formulated as an instance of DDO with a known FGM. 
To see this, let us denote the horizon of a given RL problem by $H$.
The goal is to maximize the expected value of the \textit{return} $R(\rs_{1}, \ra_{1}, \dots, \rs_{H}, \ra_{H})$, which decomposes as $\sum_{t=1}^{H}r(\rs_t, \ra_t)$.
Hence, the FGM of the return consists of edges $(\rs_t, \ra_t)$, which also form its cliques.
In this specific setting, however, we cannot control all variables directly, since an RL agent can only optimize its actions, which in turn affect the stochastic states. 
In this paper, we focus our attention on problems where all variables are optimizable, 
but we see the setting of partially-controllable variables as an exciting avenue for future work.

\vspace{-5pt}
\section{GRAPH DISCOVERY}\label{sec:discovery}
\vspace{-5pt}

{ So far, we've shown how functions with known FGMs can be optimized effectively and under much milder coverage conditions than na\"{i}ve DDO. However, in many cases, we do not know the FGM for a given function. Next, we'll discuss how to construct FGMs only from offline data practically, provided that the data satisfies a Gaussian distribution assumption:
}


\begin{restatable}{assumption}{normality}
\label{as:normal}
The inputs in the offline data follow the standard Gaussian distribution, $\rvx\sim \mathcal{N}(0, I)$.
\end{restatable}
This assumption may appear strong. However, whitening to remove correlations is a common procedure in machine learning, and in cases where the data distribution has higher order moments, we can use more advanced representation learning methods, such as variational auto-encoders \citep[VAE]{kingma2013auto, rezende2015variational, dai2019diagnosing}. Indeed, these methods are designed specifically to acquire latent representations of the data that follow the standard Gaussian distribution. 



To derive our graph discovery method, we start with the following lemma, known as the \textit{second-order Stein's identity} \citep{stein2004use, erdogdu2015newton}:
\begin{restatable}{lemma}{hessian}
   Let $\bar{\rvx}\in\mathbb{R}^{d_{\rvx}}$ and $i\neq j \in [d_{\rvx}]$. Then
    \begin{align}
        \frac{\partial^2}{\partial \bar{\rx}_j \bar{\rx}_i}\E_{\rvx\sim \Ncal}\big[f(\bar{\rvx}+\rvx)\big]=\E_{\rvx\sim \Ncal}[\rx_i \rx_j f(\bar{\rvx}+\rvx)].\nonumber
    \end{align}
\end{restatable}
Below, we demonstrate how this lemma can be used to discover FGMs from data in combination with the third criterion (in Lemma \ref{lemma:criteria}) for functional independence.
Suppose that $\rx_i$ and $\rx_j$ are not linked in a FGM of $f$, such that $(i, j)\notin \mathcal{E}$. By that criterion, 
\begin{align}
    &\frac{\partial^2}{\partial \bar{\rx}_j \partial \bar{\rx}_i}\E_{\rvx\sim \Ncal}\big[f(\bar{\rvx}+\rvx)\big] \nonumber\\
    &= \E_{\rvx\sim \Ncal}\Big[\frac{\partial^2 f}{\partial \bar{\rx}_j \partial \bar{\rx}_i}(\bar{\rvx}+\rvx)\Big] = \E_{\rvx\sim \Ncal}[0] = 0.\nonumber
\end{align}
Letting $\bar{\rvx}=0_{d_{\rvx}}$, for $(i, j)\notin \mathcal{E}$,
we have $\E_{\rvx \sim \Ncal}\big[\rx_i \rx_j f(\rvx) \big]=0.$ \footnote{Note that while it's uncertain whether the reverse claims hold in general, this graph discovery method remains practical for providing a suitable graph for our downstream tasks, i.e., DDO. }
Hence, to reconstruct edges in the unknown FGM of $f$,
one can take $M$ samples $\{\rvx^{(k)}, \ry^{(k)}\}_{k=1}^{M}$ and estimate
\begin{align}
    \label{eq:hessian}
    \hat{\rmH}_{ij} = \frac{1}{M}\sum_{k=1}^{M} \rx^{(k)}_i \rx^{(k)}_j \ry^{(k)},
\end{align}
and infer
that an edge is absent if $\hat{\rmH}_{ij}$ is close to zero. 
In our experiments, we found that a good implementation of the notion of closeness is via a Gaussian hypothesis test, $|\hat{\rmH}_{ij}|<c_{\alpha/2}/\sqrt{M}$, where $c_{\alpha/2}$ is the top-$\alpha/2$ quantile of the standard Gaussian distribution.

\section{EXPERIMENTS}
\label{sec:experiments}

\vspace{-5pt}

In this section, we present numerical experiments to verify our theoretical findings and examine how readily applicable they are in practice. We start with the setting where the FGM is known, and then explore FGM discovery methods with learned latent spaces.

\begin{figure}[t!]
\begin{center}
    \begin{subfigure}{0.45\linewidth}
    \includegraphics[width=\linewidth]{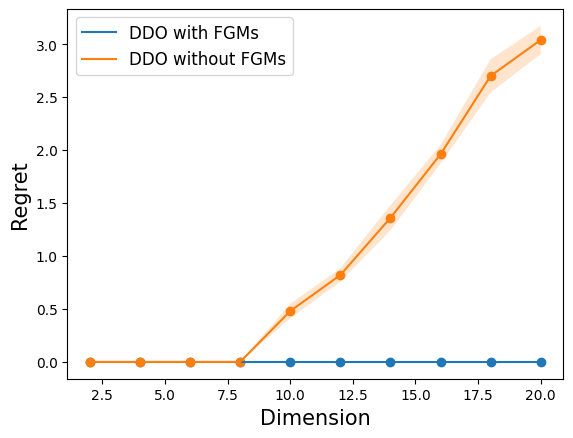}
        \caption{
        Regret (lower-better) 
    }
    \label{fig:experiment}
    \end{subfigure}
    \begin{subfigure}{0.45\linewidth}
    \includegraphics[width=\linewidth]{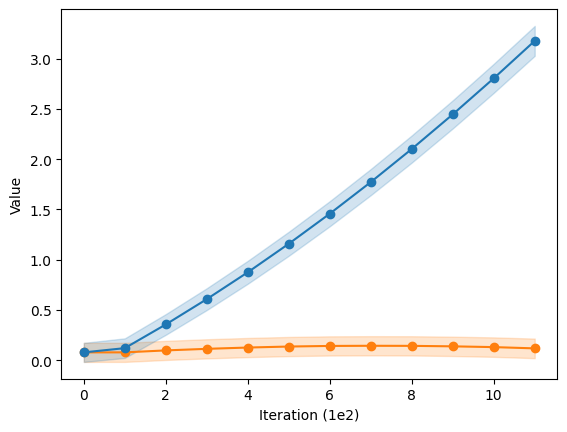}
        \caption{
        Value (higher-better)
        }
        \label{fig:experiment2}
    \end{subfigure}
\end{center}
\caption{ {\small Figure (\ref{fig:experiment}): regret (lower-better) of DDO with (\color{NavyBlue}blue\color{black}) and without (\color{Peach}orange\color{black}) FGM, against dimension in the toy quadratic problem. Averaged over 50 runs, showing 95$\%$-confidence intervals. Figure (\ref{fig:experiment2}): value (higher-better) of DDO with MLP neural networks, for Gaussian data, with (\color{NavyBlue}blue\color{black}) and without (\color{Peach}orange\color{black}) FGM, against the iteration of gradient ascent. Averaged over 128 designs, showing one-tenth of standard deviation.}}
\vspace{-15pt}
\end{figure}

\paragraph{DDO with known FGMs.}
To empirically validate the regret results from Section \ref{sec:regret}, we conducted a numerical experiment with a quadratic ground-truth function $f(\rvx)=\rx_1 \rx_2+\rx_2 \rx_3+\cdots +\rx_{d-1}\rx_d+ \rx_{d} \rx_1$, where $\forall i;\Xcal_i=\{0,1\}$, and the approximation $\hat{f}(\rvx)$ is a learned quadratic function.
We generate the data from a uniform distribution and compare na\"ive DDO in Eq. (\ref{eq:simple}) with DDO using FGMs in Eq. (\ref{eq:simple2}).
Figure~\ref{fig:experiment} shows that, as predicted by Theorem \ref{thm:regret}, the regret of DDO with FGMs remains near zero even as the problem dimensionality increases, while the regret of na\"{i}ve DDO explodes.
We then ran a more complex version of this experiment, which we visualize in Figure \ref{fig:experiment2}. 
Here, the data is drawn from the standard Gaussian distribution and the ground-truth function
is a random mixture of radial-basis functions (which we describe in more detail below). 
We fitted \emph{1)} a neural network, and \emph{2)} a neural network with FGM decomposition (see Appendix \ref{appendix:experiments} for details), and then optimized the input into each of them with gradient ascent to obtain an estimate of the optimal $\rx$. We plot the 
value of these solutions (standardized to have zero mean and unit variance), since the regret is not analytically tractable.
The results likewise show a large improvement in performance from utilizing FGMs.

\vspace{-10pt}
\paragraph{DDO with unknown FGMs.}
Next, we study the effectiveness of our algorithms in several synthetic examples where the FGM is not provided a priori, and where functional independencies must be discovered automatically, including in high-dimensional domains.
The data is generated as follows. 
We construct the ground-truth function $f(\rvx)$ by, first,
setting the set of maximal cliques $\mathcal{C}$ of its FGM to be triangles $\{0, 1, 2\}, \{2, 3, 4\}, \{4, 5, 6\}, \dots, \{d-2, d-1, d\}$. 
Crucially, these cliques overlap, introducing dependencies between their variables.
See Figure \ref{fig:example} for an illustration of the case with $d=7$.
The target function $f(\rvx)$ is a mixture of radial-basis functions,
\begin{align}
    \vspace{-5pt}
    f(\rvx) = \sum_{C\in\mathcal{C}} \rvw_C \cdot \exp(-||\rvx_C - \vmu_C||^2),\nonumber
\end{align}
where the centers $\{ \vmu_C \ | \ C\in\hat{\mathcal{C}} \}$ are sampled from a Gaussian distribution and the positive weights $\{\rvw_C \ | \ C\in\hat{\mathcal{C}}\}$ are also random. 
We evaluate $\ry=f(\rvx)$ on $N$ \textit{ground-truth} inputs drawn from the standard Gaussian distribution, $\rvx\sim \Ncal(0_{d}, I_{d})$. 
We then replace the ground-truth Gaussian inputs $\rvx$ with \emph{observable} designs $\rvx_{\text{obs}}\in\mathbb{R}^{d_{\text{obs}}}$ obtained by \textit{bijectively} composing a random affine map with the \textit{softplus} nonlinearity. 
With a slight abuse of notation, the bijection assures that the reparameterization $\ry=f(\rvx_{\text{obs}})$ is well-defined.
As such, the resulting DDO dataset takes the form $\{\rvx^{(i)}_{\text{obs}}, \ry^{(i)}\}_{i=1}^{N}$, with designs $\rvx_{\text{obs}}$ following a non-Gaussian distribution and laying in a subset of $\mathbb{R}^{d_{\text{obs}}}$.

It is important to note that, since the image of the softplus is the set of positive real numbers, the \textit{valid} values of $\rvx_{\text{obs}}$ (those for which there exists a ground-truth $\rvx$) form only a subset of $\mathbb{R}^{d_{\text{obs}}}$.
Thus, one way for a DDO method to fail is to generate a design $\rvx_{\text{obs}}$ for which there exsits no $\rvx(\rvx_{\text{obs}})$.

\begin{figure}[t!]
\begin{center}
     \begin{subfigure}{0.49\linewidth}
        \includegraphics[width=\linewidth]{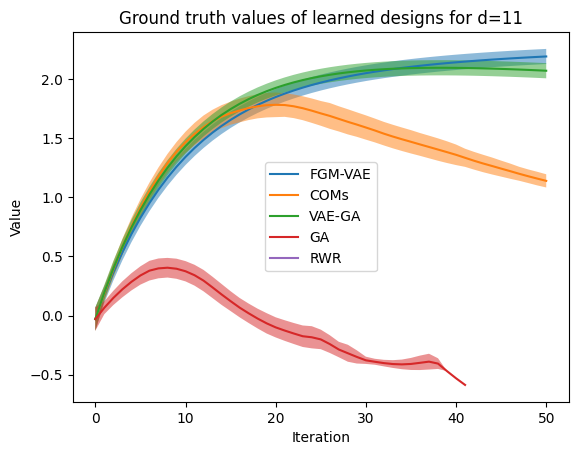}
    \end{subfigure}
    \begin{subfigure}{0.49\linewidth}
        \includegraphics[width=\linewidth]{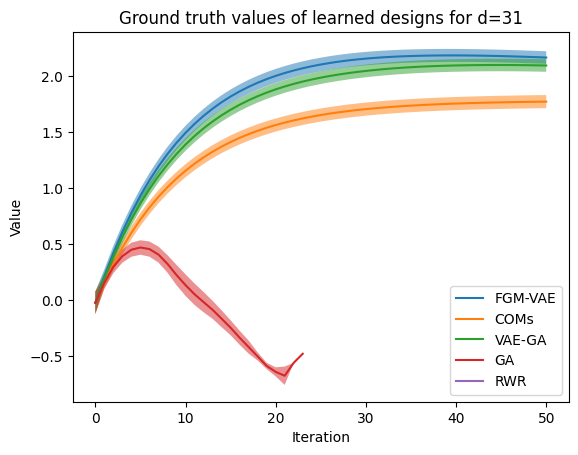}
    \end{subfigure}
    \\
    \begin{subfigure}{0.49\linewidth}
        \includegraphics[width=\linewidth]{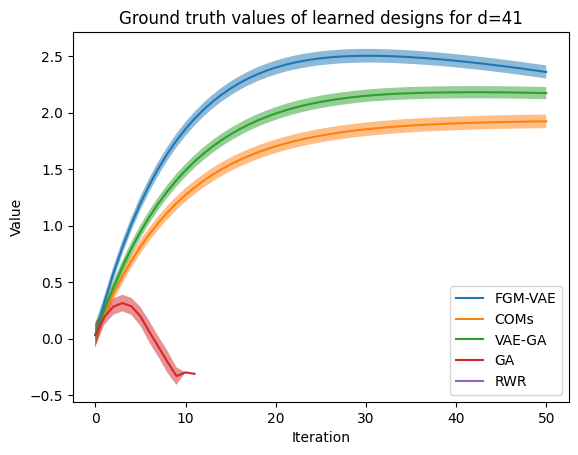}
    \end{subfigure}
    \begin{subfigure}{0.49\linewidth}
        \includegraphics[width=\linewidth]{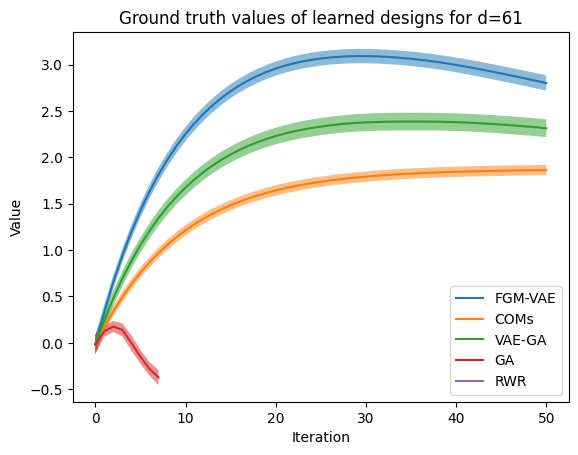}
    \end{subfigure}
\end{center}
    \caption{Values of $f(\hat{\rvx})$,
    where $\hat{\rvx}\sim \pi$, along the course of gradient ascent on $\pi$, for 11-, 21-, 31-, and 41-dimensional problems {($x$-axis: iterations, $y$-axis: design values)}. The evaluation is over the top-128 designs from a sample of 1024. Lack of curve indicates generation of invalid designs. As the dimensionality grows, the FGM decomposition becomes more ciritical.} 
    \label{fig:chain-plots}
    \vspace{-10pt}
\end{figure}

Since we do not assume access to the function's FGM, nor Gaussianity of the data, we begin by learning marginally-Gaussian representations $\rvz(\rvx_{\text{obs}})$ of the designs.
In our experiments, we choose to learn them with \textit{Variational Auto-Encoders} \citep[VAE]{kingma2013auto}.
Viewing the design values as a function of the Gaussian representations, $\ry=f(\rvz)$,
we then compute the estimator from Eq. (\ref{eq:hessian}) to infer the FGM of $f(\rvz)$ and extract its clique set $\hat{\mathcal{C}}$.
We then learn a predictive neural network model that decomposes over these cliques, and perform gradient ascent with respect to it over the $\rvz$ variable. 
Then, we decode the representation of such a design $\hat{\rvz}$ with our VAE decoder to propose a new design candidate $\hat{\rvx}_{\text{obs}}$. 
We summarize the entire algorithm for DDO with FGMs in Algorithm \ref{algorithm:gddo}.

\begin{algorithm}
\caption{DDO with FGMs}
\label{algorithm:gddo}
\begin{algorithmic}[1]
\STATE \textbf{Input}: dataset $\mathcal{D}=\{\rvx^{(i)}, \ry^{(i)} \}_{i=1}^{N}$
\STATE \color{Periwinkle}{\textbf{Preprocess} $\{\rvx^{(i)}\}_{i=1}^{N}$ so that they approximately follow the standard Gaussian distribution.}\color{black}
\STATE Approximate the cliques of the FGM of $f(\rvx)$ with Equation (\ref{eq:hessian})
\STATE Learn the functions $\hat{f}_{C}(\rvx_C)$ by minimizing the mean squared error
\begin{center}
    $\E_{(\rvx, \ry)\sim p}\big[\big( \sum_{C\in\hat{\mathcal{C}}} \hat{f}_{C}(\rvx_C) - \ry\big)^2 \big]$
\end{center}
\STATE $\hat{\pi} = \argmax_{\pi}\E_{\rvx\sim \pi}\big[ \sum_{C\in\hat{\mathcal{C}}} \hat{f}_{C}(\rvx_C)\big]$.
\STATE \color{Periwinkle}{\textbf{Invert} the preprocessing step for $\rvx\sim \hat{\pi}$.}\color{black}
\end{algorithmic}
\end{algorithm}

We compare this method with na\"{i}ve gradient ascent (GA), reward-weighted regression \citep[RWR]{peters2007reinforcement}---to see the effectiveness of na\"{i}ve methods in the tasks we designed; conservative objective models \citep[COMs]{trabucco2021conservative}---to compare to an off-the-shelf state-of-the-art method; and an ablative method that learns VAE representations and performs gradient ascent directly on them (VAE-GA), similarly to \citep{gomez2018automatic}---to isolate the benefits of FGM decomposition on top of the learned latent space. 

Results in Figure \ref{fig:chain-plots} show that incorporating FGMs into DDO leads to better designs compared to baselines, and the gap increases with the dimensionality of the problem, as expected from Theorem \ref{thm:regret}. It is also worth noting that the na\"{i}ve baselines---GA and RWR---tend to generate invalid designs quite quickly.

\begin{figure}
  \begin{center}
    \begin{subfigure}{0.49\linewidth}
        \includegraphics[width=\linewidth]{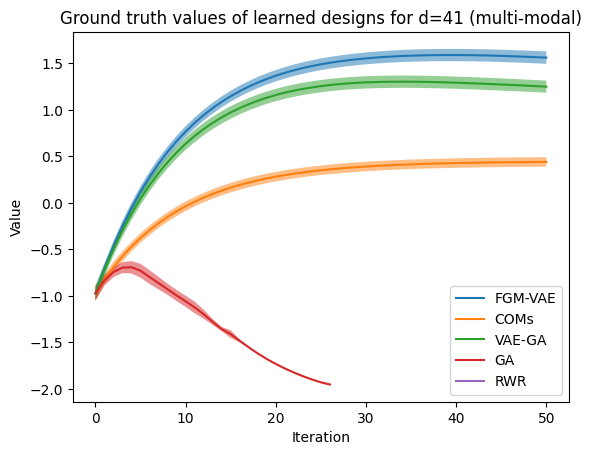}
    \end{subfigure}
    \begin{subfigure}{0.49\linewidth}
        \includegraphics[width=\linewidth]{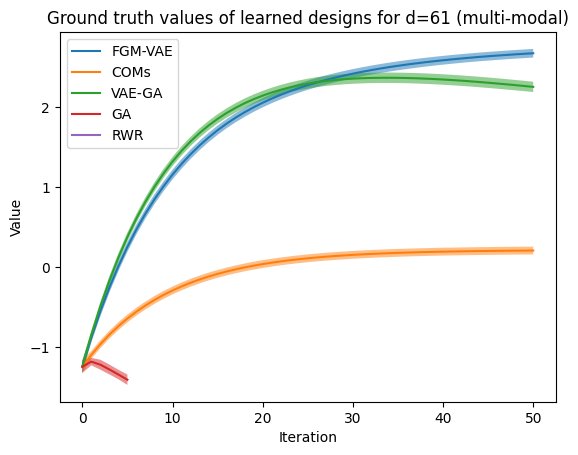}
    \end{subfigure}
  \end{center}
  \vspace{-5pt}
  \caption{Values of $f(\hat{\rvx})$,
  where $\hat{\rvx}\sim \pi$, along the course of gradient ascent on $\pi$, for 41-, and 61-dimensional problems {($x$-axis: iterations, $y$-axis: design values)}. The evaluation is over the top-128 designs from a sample of 1024. Lack of curve indicates generation of invalid designs. The unobserved base distribution generating the data was an even mixture of two Gaussian distributions $\mathcal{N}(-1, I)$ and $\mathcal{N}(1, I)$.}
  \label{fig:multimodal-chain-plots}
  \vspace{-20pt}
\end{figure}

One may wonder if Gaussianity of both the distribution of the ground-truth inputs $\rvx$ and the VAE prior simplifies the task for our algorithm, since the VAE preprocessing step  could learn to invert our non-linear data transformation.
This, however, is very unlikely given that there are infinitely many bijective reparameterizations of $\rvx$ that could explain $f(\rvx)$ perfectly and be also Gaussian (\textit{e.g.}, rotations of $\rvx$).
Nevertheless, as a sanity check, we re-ran the experiments with the ground-truth inputs $\rvx$ sampled from a mixture of Gaussians, rather than the unimodal Gaussian distribution.
This way we made sure that the representations learned by the VAE are significantly distinct from the ground-truth inputs, as well as made the multi-modality of the observed designs $\rvx_{\text{obs}}$ more explicit. 
Figure \ref{fig:multimodal-chain-plots} shows that our findings are preserved under this setting.

\vspace{-10pt}
\section{CONCLUSION}

We showed how DDO can be made feasible and tractable by incorporating graphical factorization structure.
We capture this structure with functional graphical models (FGMs), and we prove that a function can decompose over its maximal cliques analogously to how probabilistic graphical models (PGMs) decompose probability distributions. We then showed theoretically that DDO with FGMs admits sample-efficient optimization without requiring that the data distribution to directly cover optimal designs, instead only requiring coverage of optimal values for individual cliques of the FGM.
While we did verify our findings experimentally, 
our experiments serve mainly as numerical verification of our theoretical findings.
Enabling data-efficient optimization in practical real-world settings will require additional development of our proposed algorithmic framework, and presents a number of exciting challenges for future work. For example, while our proposed FGM discovery approach is relatively simple and pragmatic, the general problem of discovering such functional independence relationships might relate to other fields that are concerned with recovery structure: discovery of probabilistic independence relationships, causal inference, and structure learning. Incorporating ideas from these fields into the FGM framework might not only constitute an exciting direction for future work, but might also lead to powerful new methods that can discover significantly improved designs from suboptimal datasets.

\section*{Acknowledgements}
We thank Seohong Park for inspiring discussions on technical aspects of FGMs and James Bowden for inspiring conversations on their applications to scientific problems.

\bibliography{main.bib}
\bibliographystyle{chicago}

\clearpage
 
\appendix 
\onecolumn

\section{Auxiliary Facts}
\label{appendix:auxiliary}
\begin{restatable}[Hammersley-Clifford \citep{clifford1971markov}]{theorem}{hc}
    Let $\mathcal{G}=(\mathcal{V}, \mathcal{E})$ be a simple graph with clique set $\mathcal{C}$, and let $\rvx = \rvx_{\mathcal{V}} \sim p$, where $p(\rvx)$ is a strictly positive probability distribution.  If $p$ is Markov with respect to $\mathcal{G}$, then, $\forall \rvx\in\mathcal{X}$,
    \begin{align}
        p(\rvx) = \prod_{C\in\mathcal{C}} \phi_{C}(\rvx_{C}),\nonumber
    \end{align}
    for some functions $\phi_{C}(\rvx_{C})>0$.
\end{restatable}

\hessian*
\begin{proof}
We start from proving a known result. Namely,
\begin{align}
    \label{eq:gaussian-derivative}
        \frac{\partial}{\partial \bar{\rx}_i}\E_{\rvx\sim \Ncal}\big[f(\bar{\rvx}+\rvx)]=\E_{\rvx\sim \Ncal}[\rx_i f(\bar{\rvx}+\rvx)].
    \end{align}
To do it, we use integration by parts,
\begin{align}
    \frac{\partial}{\partial \bar{\rx}_i}\E_{\rvx\sim \Ncal}[f(\bar{\rvx}+\rvx)] &= \E_{\rvx\sim \Ncal}\Big[\frac{\partial}{\partial \bar{\rx}_i}f(\bar{\rvx}+\rvx)\Big]\nonumber\\
    &= \E_{\rvx\sim \Ncal}\Big[\frac{\partial}{\partial \rx_i}f(\bar{\rvx}+\rvx)\Big]\nonumber\\
    &= \E_{\rvx_{-i}\sim \Ncal}\Big[ \int_{\rx_i} d\rx_i \frac{1}{\sqrt{2\pi}}\exp\Big( \frac{-1}{2}\rx^{2}_i\Big) \frac{\partial}{\partial \rx_i}f(\bar{\rvx}+\rvx) \Big]\nonumber\\
    &= \E_{\rvx_{-i}\sim \Ncal}\Big[ \Big[ \frac{1}{\sqrt{2\pi}}\exp\Big( \frac{-1}{2}\rx^{2}_i\Big) f(\bar{\rvx}+\rvx) \Big]^{\infty}_{-\infty} + \int_{\rx_i}d\rx_i\frac{\rx_i}{\sqrt{2\pi}}\exp\Big( \frac{-1}{2}\rx^{2}_i\Big) f(\bar{\rvx}+\rvx)  \Big] \nonumber\\
    &= \E_{\rvx_{-i}\sim \Ncal}\Big[ 0 + \int_{\rx_i}d\rx_i\frac{\rx_i}{\sqrt{2\pi}}\exp\big( \frac{-1}{2}\rx^{2}_i\big) f(\bar{\rvx}+\rvx)  \Big] \nonumber\\
    &= \E_{\rvx\sim \Ncal}[\rx_i f(\bar{\rvx}+\rvx)].\nonumber
\end{align}
Now, applying this result twice,
\begin{align}
    \frac{\partial^2}{\partial \bar{\rx}_j \bar{\rx}_i}\E_{\rvx\sim \Ncal}\big[f(\bar{\rvx}+\rvx)] &= \frac{\partial}{\partial \bar{\rx}_j}\frac{\partial}{\partial \bar{\rx}_i}\E_{\rvx\sim \Ncal}[f(\bar{\rvx}+\rvx)]\nonumber\\
    &= \frac{\partial}{\partial \bar{\rx}_j}\E_{\rvx\sim \Ncal}[\rx_i f(\bar{\rvx}+\rvx)]=\E_{\rvx\sim \Ncal}[\rx_i \rx_j f(\bar{\rvx}+\rvx)],\nonumber
\end{align}
as required.
\end{proof}

\clearpage
\section{Omitted Proofs Of Our Results}\label{ape:proof}

\subsection{Proof of Lemma~\ref{lemma:criteria} } 

\label{appendix:ours}
\difindcond*
\begin{proof}
    1$\iff$2: Let us, for clarity, write $\underline{A}=A\setminus S$ and $\underline{B}=B\setminus S$. We suppose that 1 holds as in Definition \ref{def:difind},
    \begin{align}
        \label{eq:dd-rewritten}
        \frac{\partial f}{\partial \rvx_{\underline{A}}} (\rvx) = F_{-\underline{B}}(\rvx_{\mathcal{V}\setminus \underline{B}}).
    \end{align}
    Integrating it with respect to $\rvx_{\underline{A}}$ gives
    \begin{align}
        f(\rvx) 
        &= \int_{\rvx_{\underline{A}}} F_{-\underline{B}}(\rvx_{\mathcal{V}\setminus \underline{B}})d \rvx_{\underline{A}} + f_{-\underline{A}}(\rvx_{\mathcal{V}\setminus \underline{A}})\nonumber\\
        &= f_{-\underline{B}}(\rvx_{\mathcal{V}\setminus \underline{B}}) + f_{-\underline{A}}(\rvx_{\mathcal{V}\setminus \underline{A}}),\nonumber
    \end{align}
    for some $f_{-\underline{B}}$ and $f_{-\underline{A}}$. This proves statement 2 and allows us to recover Equation (\ref{eq:dd-rewritten}) for $\underline{A}$ ($\underline{B}$) by differentiating with respect to $\rvx_{\underline{A}}$ ($\rvx_{\underline{B}}$).

    1$\iff$3: Suppose that 1 holds as in Definition \ref{def:difind}. Since the right-hand side of the equation is not a function of $\rvx_{\underline{B}}$, differentiating the equation with respect to $\rvx_{\underline{B}}$ gives $0$, which proves statement 3. We recover statement 1 from statement 3 by integrating with respect to $\rvx_{\underline{B}}$ (or $\rvx_{\underline{A}}$).
\end{proof}

\subsection{Proof of Theorem \ref{theorem:decomp}}
\decomp*
\begin{proof}
    Consider a probability distribution defined by
\begin{align}
    e_{f}(\rvx) = \frac{\exp f(\rvx)}{Z}, \ \ \text{where} \ \ Z\triangleq \int_{\rvx} \exp f(\rvx) d \rvx.\nonumber
\end{align}
Then, let $i$ and $j$ be nodes in $\mathcal{G}$ that are not linked: $(i, j)\notin \mathcal{E}$. By Definition \ref{def:sg}, it follows that 
\begin{align}
    e_{f}(\rvx) &= \frac{ \exp \big(f_{-i}(\rvx_{-i}) + f_{-j}(\rvx_{-j}) \big) }{Z} \nonumber\\
    &= \frac{ \exp f_{-i}(\rvx_{-i}) }{\sqrt{Z}} \frac{\exp f_{-j}(\rvx_{-j})}{\sqrt{Z}}.\nonumber
\end{align}
That is, measure $e_{f}(\rvx)$ factorizes over $\{\mathcal{V}\setminus\{i\}, \mathcal{V}\setminus\{j\}\}$. 
Hence, random variables $\rx_i$ and $\rx_j$ are (probabilistically) independent given $\rvx_{\mathcal{V}\setminus \{i, j\}}$ \cite{studeny2006probabilistic}. As the pair $(i, j)$ was arbitrary, this implies that $e_{f}(\rvx)$ satisfies the \textit{pairwise Markov property} with respect to $\mathcal{G}$ (equivalently, $\mathcal{G}$ is an I-map of $e_f(\rvx)$). As the measure induced by $e_{f}(\rvx)$ is strictly positive, \textit{Hammersley-Clifford Theorem} \cite{clifford1971markov} implies that $e_{f}(\rvx)$ factorizes according to $\mathcal{G}$. That is, for the clique set $\mathcal{C}$ of $\mathcal{G}$, there exist positive functions $\{u_{C}(\rvx_{C}) \ | \  C\in\mathcal{C}\}$, such that 
\begin{align}
    e_{f}(\rvx) = \prod_{C\in\mathcal{C}} u_{C}(\rvx_C).\nonumber
\end{align}

Taking the logarithm on both sides, we get
\begin{align}
    f(\rvx) = \sum_{C\in\mathcal{C}} \log u_{C}(\rvx_{C}) + \log Z
    \triangleq \sum_{C\in\mathcal{C}} f_{C}(\rvx_{C}) + \log Z.\nonumber
\end{align}
That is, the original function $f(\rvx)$ can be expressed as a sum of functions of cliques of graph $\mathcal{G}$ (up to an input-independent constant). Of course, the constant can be subsumed into the decomposing functions. This lets us conclude the proof of the theorem. 

\end{proof}
\subsection{Proof of Theorem~\ref{thm:regret}}
\label{subsec:regret}
\regretfgm*

\begin{proof}
Following the standard literature of regression \cite{agarwal2019reinforcement} and using Assumption (3), by leveraging $f_C \in \Fcal_C$, we can obtain with probability $1-\delta$, 
\begin{align}
    \E_{\rvx\sim p}\Big[\big(\hat f(\rvx) - f(\rvx) \big)^2\Big] \leq \mathrm{Stat},\quad \mathrm{Stat}=D \times  \frac{\sum_{C\in \mathcal{C}} \log (|\Fcal_C|/\delta)}{n}.
\end{align}
where $\hat f(\rvx)=\sum_C \hat f_C(\rvx_C),f(\rvx)=\sum_C f_C(\rvx_C)$, and $D$ is universal constant, noting $
   \log( |\Fcal|)=\sum_C \log(|\Fcal_C|). $
Note this statement states that the MSE guarantee is ensured for the whole $\hat f(\rvx)$. 

Now, we have 
\begin{align*}
    & \E_{\rvx\sim p}\big[\big(\hat f(\rvx) - f(\rvx) \big)^2\big] \\ 
 & = \EE_{\rvx\sim p}\Bigg[ \Big(\sum_C  \{\hat f_C(\rvx_C)  -  f_C(\rvx_C)\} \Big)^2 \Bigg] \\ 
 &= \sum_{C}\E_{\rvx\sim p}\left [  \{\hat f_C(\rvx_C)  -  f_C(\rvx_C)\}^2 \right ] + \sum_{C_1 \neq C_2}\E_{\rvx\sim p}\left [  \{\hat f_{C_1}(\rvx_{C_1})  -  f_{C_1}(\rvx_{C_1})\} \{\hat f_{C_2}(\rvx_{C_2})  -  f_{C_2}(\rvx_{C_2})\} \right ].  
\end{align*}
Here, using Assumption (1) and (2), note 
\begin{align*}
    &  \E_{\rvx\sim p}\left [  \{\hat f_{C_1}(\rvx_{C_1})  -  f_{C_1}(\rvx_{C_1})\} \{\hat f_{C_2}(\rvx_{C_2})  -  f_{C_2}(\rvx_{C_2})\} \right ]\\
    &\leq  \sigma \times \left\{\EE_{\rvx\sim p}\left [ \{ \hat f_{C_1}(\rvx_{C_1})-  f_{C_1}(\rvx_{C_1})\}^2 \right]^{1/2} \right\}\times \left\{ \EE_{\rvx\sim p}\left[ \left\{\hat f_{C_2}(\rvx_{C_2}) -  f_{C_2}(\rvx_{C_2})\right\}^2 \right ]^{1/2} \right\}. 
\end{align*}
Hence, we get 
\begin{align}\label{eq:decomposition}
&\EE_{\rvx\sim p}[\{\hat f(\rvx) - f(\rvx) \}^2] \\
 &\geq \sum_{C}\EE_{\rvx\sim p}\left [  \{\hat f_C(\rvx_C)  -  f_C(\rvx_C)\}^2 \right ] \nonumber\\
 &+ \sigma \times \left\{\EE_{\rvx\sim p}\left [ \{ \hat f_{C_1}(\rvx_{C_1})-  f_{C_1}(\rvx_{C_1})\}^2 \right]^{1/2} \right\}\times \left\{ \EE_{\rvx\sim p}\left[ \left\{\hat f_{C_2}(\rvx_{C_2}) -  f_{C_2}(\rvx_{C_2})\right\}^2 \right ]^{1/2} \right\}\\
&\geq  \sigma \left\{\sum_C  \EE_{\rvx\sim p}\left [ \{ \hat f_{C}(\rvx_{C})-  f_{C}(\rvx_{C})\}^2 \right]^{1/2}  \right \}^2   + \{1-\sigma\} \sum_{C \in \mathcal{C}}\EE_{\rvx\sim p}\left [  \{\hat f_C(\rvx_C)  -  f_C(\rvx_C)\}^2 \right ]. \\
&\geq \{1-\sigma\} \sum_{C \in \mathcal{C}}\EE_{\rvx\sim p}\left [  \{\hat f_C(\rvx_C)  -  f_C(\rvx_C)\}^2 \right ]. 
\end{align}
Therefore, we have 
\begin{align}
    \label{ineq:stat}
   \sum_{C \in \mathcal{C}}\EE_{\rvx\sim p}[\{\hat f_C(\rvx_C)  -  f_C(\rvx_C)\}^2 ]\leq \frac{1}{1-\sigma}\times \mathrm{Stat}.  
\end{align}
Then, for any policy $\pi \in \Pi$, we get  
\begin{align*}
    J(\pi) - J(\hat \pi) & \leq  J(\pi) - \EE_{\rvx \sim \pi}[\hat f(\rvx)] + \EE_{\rvx\sim \pi}[\hat f(\rvx)] - \EE_{\rvx\sim \hat \pi}[\hat f(\rvx)] + \EE_{\rvx\sim \hat \pi}[\hat f(\rvx)] - J(\hat \pi) \\ 
    &= J(\pi) - \EE_{\rvx\sim \pi}[\hat f(\rvx)] + \EE_{\rvx \sim \hat \pi}[\hat f(\rvx)] - J(\hat \pi)\\
    & = \EE_{\rvx\sim \pi}[f(\rvx)] -  \EE_{\rvx\sim \pi}[\hat f(\rvx)] +  \EE_{\rvx\sim \hat \pi}[\hat f(\rvx)] - \EE_{\rvx \sim \hat \pi}[f(\rvx)]. 
\end{align*}
We use the following to upper-bound last terms:
\begin{align*}
    &|\EE_{\rvx \sim \pi}[f(\rvx)-\hat f(\rvx)]| \leq \sum_{C \in \mathcal{C}} \EE_{\rvx \sim \pi}[| f_C(\rvx_C)-\hat f_C(\rvx_C)|] \tag{Triangle inequality}\\
  &\leq  \left\{\max_{ C \in \mathcal{C}} \max_{\rvx_C \in \mathcal{X}_C} \frac{\pi(\rvx_C)}{p(\rvx_C)}\right\} \sum_{C \in \mathcal{C}} \EE_{\rvx \sim p}[| f_C(\rvx)-\hat f_C(\rvx)|]   \tag{Importance sampling}  \\ 
   & \leq  \left\{\max_{ C \in \mathcal{C}} \max_{\rvx_C \in \mathcal{X}_C} \frac{\pi(\rvx_C)}{p(\rvx_C)}\right\} \sum_{C \in \mathcal{C}}\sqrt{\EE_{\rvx \sim p}[ \{f_C(\rvx_C)-\hat f_C(\rvx_C)\}^2]} \tag{Jensen's inequality}  \\ 
     &\leq  \left\{\max_{ C \in \mathcal{C}} \max_{\rvx_C \in \mathcal{X}_C} \frac{\pi(\rvx_C)}{p(\rvx_C)}\right\} \sqrt{|\mathcal{C}|}\sqrt{\sum_{C \in \mathcal{C}} \EE_{\rvx \sim p}[ \{f_C(\rvx_C)-\hat f_C(\rvx_C)\}^2]} \tag{CS inequality} \\ 
     &=\max_{ C \in \mathcal{C}} \max_{\rvx_C \in \mathcal{X}_C}\frac{\pi(\rvx_C)}{p(\rvx_C)} \times  \sqrt{|\mathcal{C}|}\times \sqrt{\frac{\mathrm{Stat}}{1-\sigma}}. 
      \tag{Use Ineq. \ref{ineq:stat}}
\end{align*}
Finally, combining everything, we have 
\begin{align*}
    J(\pi) - J(\hat \pi) & \leq 2\max_{\pi\in \Pi}\max_{ C \in \mathcal{C}} \max_{\rvx_C \in \mathcal{X}_C}\frac{\pi(\rvx_C)}{p(\rvx_C)} \times \sqrt{|\mathcal{C}|} \mathrm{Stat} \\
    &\lesssim \max_{\pi\in \Pi}\max_{ C \in \mathcal{C}} \max_{\rvx_C \in \mathcal{X}_C}\frac{\pi(\rvx_C)}{p(\rvx_C)}\times \sqrt{\frac{|\mathcal{C}|\sum_{C\in \mathcal{C}} \log (|\Fcal_C|/\delta)}{n}}\times \sqrt{\frac{1}{1-\sigma}}. 
\end{align*}
\end{proof}

\subsection{Proof of Lemma~\ref{lem:helpful}}
\lemmacoverage*

\begin{proof}
Let $\rvx = [\rvx^{\top}_C, \{\rvx'\}^{\top}_C]^{\top}$. Then, 
\begin{align}\label{eq:inequality}
   \max_{\rvx'_C} \frac{\pi(\rvx)}{p(\rvx)}= \frac{\pi(\rvx_C)}{p(\rvx_C)} \times 
 \max_{\rvx'_C}  \frac{\pi(\rvx'_C|\rvx_{C})}{p(\rvx'_C|\rvx_{C})}  \geq  \frac{\pi(\rvx_C)}{p(\rvx_C)}.
\end{align}

 Here, we use  $ \max_{\rvx'_C}  \frac{\pi(\rvx'_C|\rvx_{C})}{p(\rvx'_C|\rvx_{C})} \geq 1$.
 This is because if $ \max_{\rvx'_C}  \frac{\pi(\rvx'_C|\rvx_{C})}{p(\rvx'_C|\rvx_{C})} < 1$, we would get contradiction:
 \begin{align*}
 1=\sum_{\rvx'_C}\pi(\rvx'_C|\rvx_{C}) \leq  \sum_{\rvx'_C}  p(\rvx'_C|\rvx_{C}) \max_{\rvx'_C} \frac{\pi(\rvx'_C|\rvx_{C})}{p(\rvx'_C|\rvx_{C})}  <  \sum_{\rvx'_C}  p(\rvx'_C|\rvx_{C}) =1. 
 \end{align*}
Finally, using Eq.~\ref{eq:inequality}, we obtain 
 \begin{align}
     \max_{\rvx}  \frac{\pi(\rvx)}{p(\rvx)}  = \max_{\rvx_C}\max_{\rvx'_C}  \frac{\pi(\rvx)}{p(\rvx)} \geq \max_{\rvx_C} \frac{\pi(\rvx_C)}{p(\rvx_C)}. \nonumber
 \end{align}    
\end{proof}
\clearpage

\section{Details of Experiments}
\label{appendix:experiments}
We conducted our experiments in a Google Colab with TPU. One can access it via the following link:
\begin{center}
    {\small \url{https://colab.research.google.com/drive/1qt4M3C35bvjRHPIpBxE3zPc5zvX6AAU4?usp=sharing}.}
\end{center}

In all runs, all neural networks were two-hidden layer MLPs, with 256 hidden units. 
Each of them was trained with Adam \citep{kingma2014adam} optimization algorithm with batch size of 128.
Hyperparameters for COMs were taken from the COMs paper \citep{trabucco2021conservative} and we did not observe any improvement by tuning them.
As hyperparameters for all other methods are a subset of hyperparameters of COMs, we used the same hyperparameters, except that methods that use VAEs \citep{kingma2013auto}
have 10 times smaller gradient ascent learning rate.
We also used the common choice for temperature $\tau=0.05$ for reward-weighted regression \citep{peters2007reinforcement, peng2019advantage, kostrikov2021offline}.

\paragraph{To represent a decomposed function} $f(\rvx) = \sum_{C\in\mathcal{C}}f_{C}(\rvx_{C})$, we use a single neural-network model of the same class as for all other methods. In our experiments, it was a multi-layer perceptron (MLP). The decomposition was implemented by duplicating the input (row) vector $\rvx$ $|\mathcal{C}|$ times and stacking the copies horizontally. Each copy corresponded to one of the cliques. Then, for each clique we masked out inputs that were not included. That is, we multiplied the stacked copies of $\rvx$ by the matrix $\mC$ such that $\mC_{ij}=1$ if $\rvx_j$ is in the $i^{th}$ clique, and $0$ otherwise. That is, we form the following input
\begin{align}
    \begin{bmatrix}
        \mC_{11} \rvx_1 \quad \dots \quad \mC_{1d} \rvx_d\\
        \\
        \dots \\
        \\
        \mC_{|\Ccal|1} \rvx_1 \quad \dots \quad \mC_{|\Ccal|d} \rvx_d
    \end{bmatrix}
    \nonumber
\end{align}
We pass the rows to the MLP in parallel and return the mean of the $|\mathcal{C}|$ output values---this is equivalent to the summation from Theorem \ref{theorem:decomp} in terms of represtational power but we believe it may be more numerically stable when $|\mathcal{C}|\rightarrow \infty$.
\end{document}


\twocolumn[

\aistatstitle{Enabling Data-Driven Optimization With Differential Independence Graphs: Supplement}

\section{Auxiliary Facts}
\label{appendix:auxiliary}
\begin{restatable}[Hammersley-Clifford \cite{clifford1971markov}]{theorem}{hc}
    Let $\mathcal{G}=(\mathcal{V}, \mathcal{E})$ be a simple graph with clique set $\mathcal{C}$, and let $\rvx = \rvx_{\mathcal{V}} \sim p$, where $p(\rvx)$ is a strictly positive probability distribution.  If $p$ is Markov with respect to $\mathcal{G}$, then, $\forall \rvx\in\mathcal{X}$,
    \begin{align}
        p(\rvx) = \prod_{C\in\mathcal{C}} \phi_{C}(\rvx_{C}),\nonumber
    \end{align}
    for some functions $\phi_{C}(\rvx_{C})>0$.
\end{restatable}

\hessian*
\begin{proof}
We start from proving a known result. Namely,
\begin{align}
    \label{eq:gaussian-derivative}
        \frac{\partial}{\partial \bar{\rz}_i}\E_{\rvz\sim N}\big[f(\bar{\rvz}+\rvz)]=\E_{\rvz\sim N}[\rz_i f(\bar{\rvz}+\rvz)].
    \end{align}
To do it, we use integration by parts,
\begin{align}
    \frac{\partial}{\partial \bar{\rz}_i}\E_{\rvz\sim N}[f(\bar{\rvz}+\rvz)] &= \E_{\rvz\sim N}\Big[\frac{\partial}{\partial \bar{\rz}_i}f(\bar{\rvz}+\rvz)\Big]\nonumber\\
    &= \E_{\rvz\sim N}\Big[\frac{\partial}{\partial \rz_i}f(\bar{\rvz}+\rvz)\Big]\nonumber\\
    &= \E_{\rvz_{-i}\sim N}\Big[ \int_{\rz_i} d\rz_i \frac{1}{\sqrt{2\pi}}\exp\Big( \frac{-1}{2}\rz^{2}_i\Big) \frac{\partial}{\partial \rz_i}f(\bar{\rvz}+\rvz) \Big]\nonumber\\
    &= \E_{\rvz_{-i}\sim N}\Big[ \Big[ \frac{1}{\sqrt{2\pi}}\exp\Big( \frac{-1}{2}\rz^{2}_i\Big) f(\bar{\rvz}+\rvz) \Big]^{\infty}_{-\infty} + \int_{\rz_i}d\rz_i\frac{\rz_i}{\sqrt{2\pi}}\exp\Big( \frac{-1}{2}\rz^{2}_i\Big) f(\bar{\rvz}+\rvz)  \Big] \nonumber\\
    &= \E_{\rvz_{-i}\sim N}\Big[ 0 + \int_{\rz_i}d\rz_i\frac{\rz_i}{\sqrt{2\pi}}\exp\big( \frac{-1}{2}\rz^{2}_i\big) f(\bar{\rvz}+\rvz)  \Big] \nonumber\\
    &= \E_{\rvz\sim N}[\rz_i f(\bar{\rvz}+\rvz)].\nonumber
\end{align}
Now, applying this result twice,
\begin{align}
    \frac{\partial^2}{\partial \bar{\rz}_j \bar{\rz}_i}\E_{\rvz\sim N}\big[f(\bar{\rvz}+\rvz)] &= \frac{\partial}{\partial \bar{\rz}_j}\frac{\partial}{\partial \bar{\rz}_i}\E_{\rvz\sim N}[f(\bar{\rvz}+\rvz)]\nonumber\\
    &= \frac{\partial}{\partial \bar{\rz}_j}\E_{\rvz\sim N}[\rz_i f(\bar{\rvz}+\rvz)]=\E_{\rvz\sim N}[\rz_i \rz_j f(\bar{\rvz}+\rvz)],\nonumber
\end{align}
as required.
\end{proof}

\clearpage
\section{Omitted Proofs Of Our Results}
\label{appendix:ours}
\difindcond*
\begin{proof}
    1$\iff$2: Let us, for clarity, write $\underline{A}=A\setminus S$ and $\underline{B}=B\setminus S$. We suppose that 1 holds as in Definition \ref{def:difind},
    \begin{align}
        \label{eq:dd-rewritten}
        \frac{\partial f}{\partial \rvx_{\underline{A}}} (\rvx) = F_{-\underline{B}}(\rvx_{\mathcal{V}\setminus \underline{B}}).
    \end{align}
    Integrating it with respect to $\rvx_{\underline{A}}$ gives
    \begin{align}
        f(\rvx) 
        &= \int_{\rvx_{\underline{A}}} F_{-\underline{B}}(\rvx_{\mathcal{V}\setminus \underline{B}})d \rvx_{\underline{A}} + f_{-\underline{A}}(\rvx_{\mathcal{V}\setminus \underline{A}})\nonumber\\
        &= f_{-\underline{B}}(\rvx_{\mathcal{V}\setminus \underline{B}}) + f_{-\underline{A}}(\rvx_{\mathcal{V}\setminus \underline{A}}),\nonumber
    \end{align}
    for some $f_{-\underline{B}}$ and $f_{-\underline{A}}$. This proves statement 2 and allows us to recover Equation (\ref{eq:dd-rewritten}) for $\underline{A}$ ($\underline{B}$) by differentiating with respect to $\rvx_{\underline{A}}$ ($\rvx_{\underline{B}}$).

    1$\iff$3: Suppose that 1 holds as in Definition \ref{def:difind}. Since the right-hand side of the equation is not a function of $\rvx_{\underline{B}}$, differentiating the equation with respect to $\rvx_{\underline{B}}$ gives $0$, which proves statement 3. We recover statement 1 from statement 3 by integrating with respect to $\rvx_{\underline{B}}$ (or $\rvx_{\underline{A}}$).
\end{proof}

\clearpage
\section{Visualizations}
\label{appendix:viz}
\begin{figure}
    \centering
    \begin{subfigure}{0.32\textwidth}
        \includegraphics[width=\textwidth]{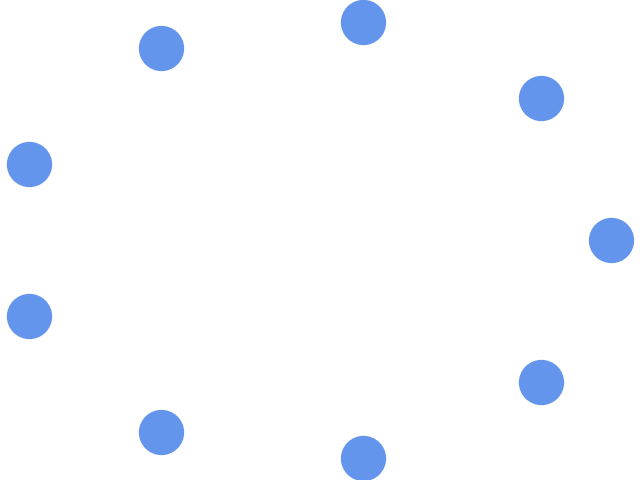}
        \caption{No edges.}
    \end{subfigure}
    \begin{subfigure}{0.34\textwidth}
        \includegraphics[width=\textwidth]{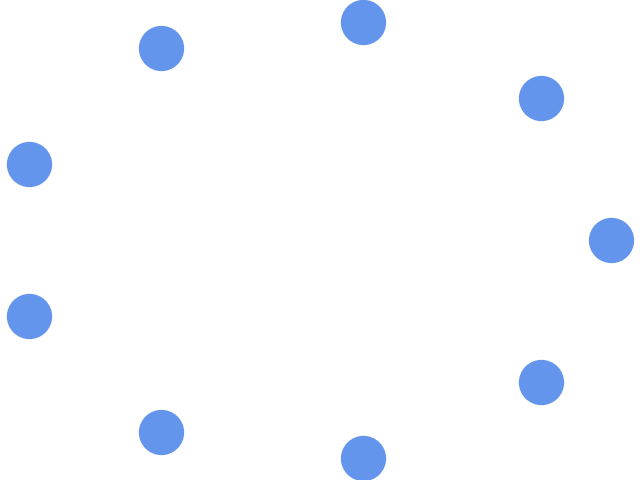}
        \caption{No edges.}
    \end{subfigure}
    \begin{subfigure}{0.32\textwidth}
        \includegraphics[width=\textwidth]{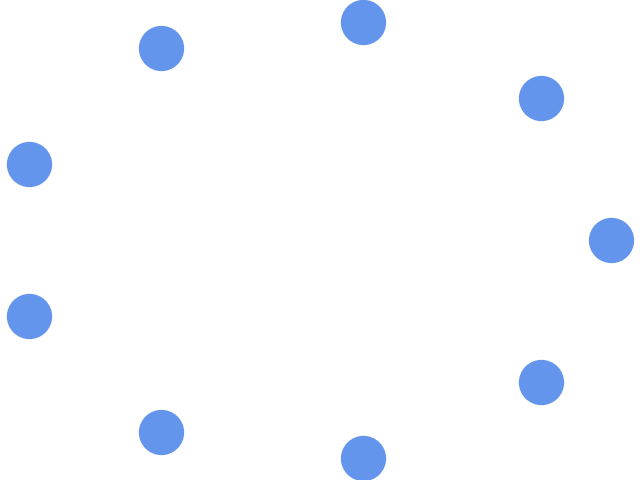}
        \caption{No edges.}
    \end{subfigure}\\
        \begin{subfigure}{0.32\textwidth}
        \includegraphics[width=\textwidth]{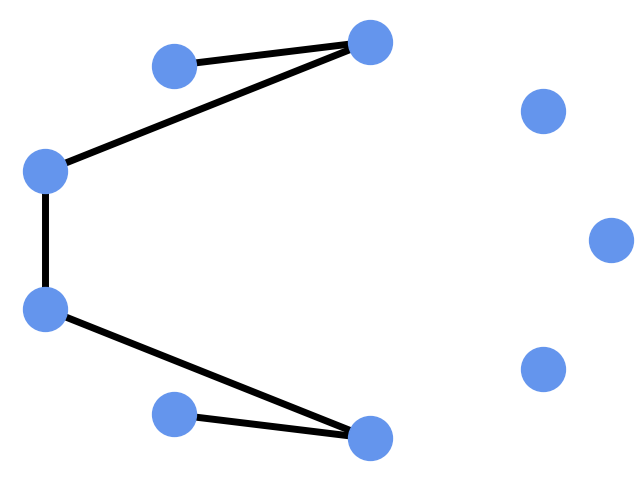}
        \caption{5 edges. Density 0.14.}
    \end{subfigure}
    \begin{subfigure}{0.34\textwidth}
        \includegraphics[width=\textwidth]{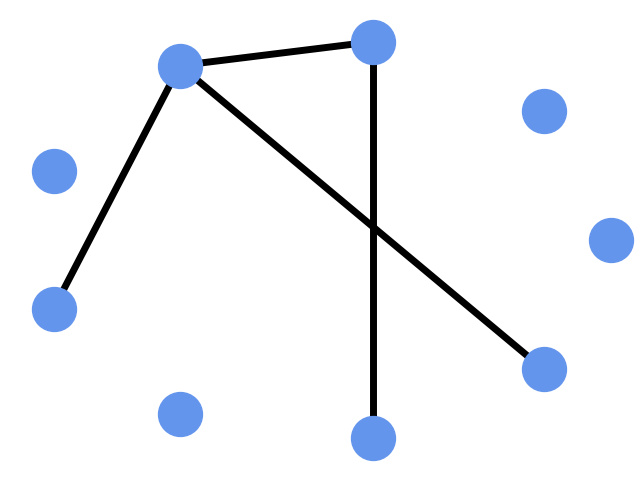}
        \caption{4 edges. Density 0.11.}
    \end{subfigure}
    \caption{Graphs discovered by GGD in SuperConductor task. As we can see, they tend to be sparse, which largely simplifies the fitted model, but does not harm performance.}
    \label{fig:superconductor-graphs}
\end{figure}

In Figure \ref{fig:superconductor-graphs}, we show graphs that Algorithm \ref{algorithm:ggd} discovered in the 9-dimensional latent space in the SuperConductor task. The discovered graphs are very sparse, often edgeless. This massively simplifies the model expressiveness, and yet does not harm the performance. In fact, in our experiments, we found that high values of $\alpha$ or the momentum parameter (both of which control sparsity) tend to give better final results.

\begin{figure}
    \centering
    \includegraphics[width=0.5\textwidth]{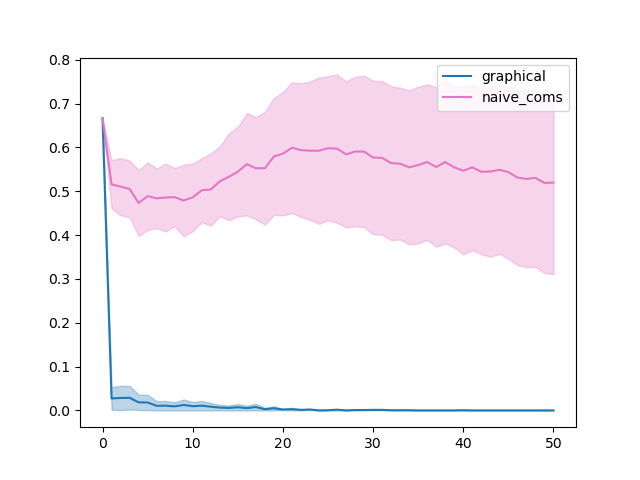}
    \caption{The fraction of invalid candidates produced by each algorithm}
    \label{fig:invalid}
\end{figure}

In Figure \ref{fig:invalid}, we plot the fraction of invalid candidates produced by each algorithm (Algorithm \ref{algorithm:gddo} and COMs) in the 3d toy task. That is, the proportion of such $\hat{\rvx}$ for which a ``groundt-truth" $\hat{\rvz}$ does not exist. While Algorithm \ref{algorithm:gddo} quickly learns the right manifold, COMs tends to produce invalid candidates.

\begin{figure}
    \centering
    \begin{subfigure}{0.32\textwidth}
        \includegraphics[width=\textwidth]{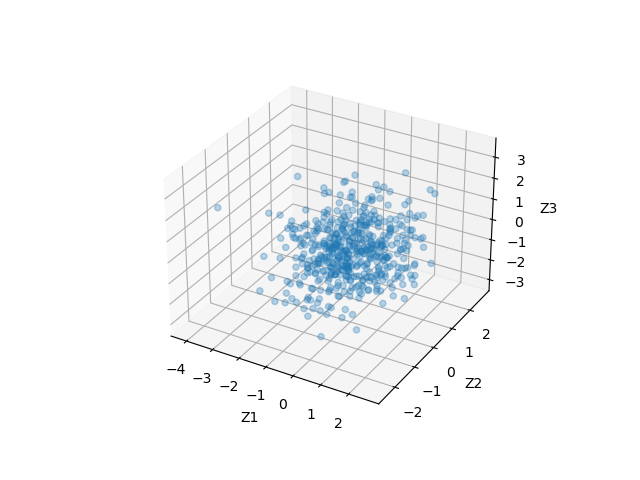}
    \end{subfigure}
    \begin{subfigure}{0.34\textwidth}
        \includegraphics[width=\textwidth]{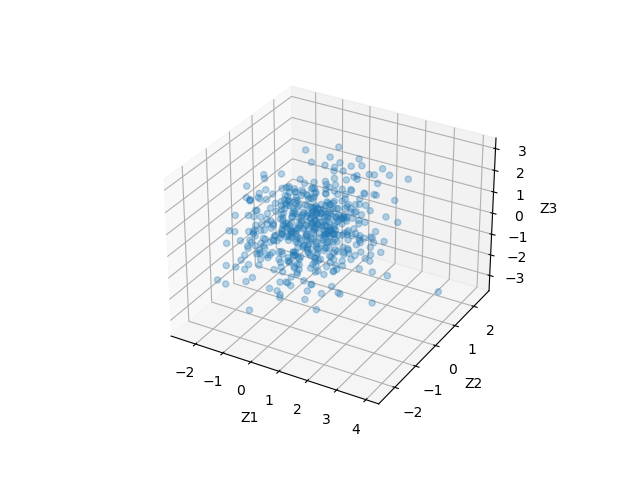}
    \end{subfigure}
    \begin{subfigure}{0.32\textwidth}
        \includegraphics[width=\textwidth]{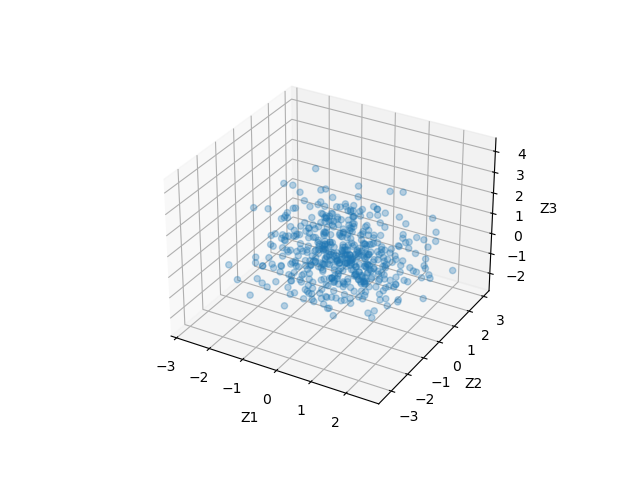}
    \end{subfigure}\\
        \begin{subfigure}{0.32\textwidth}
        \includegraphics[width=\textwidth]{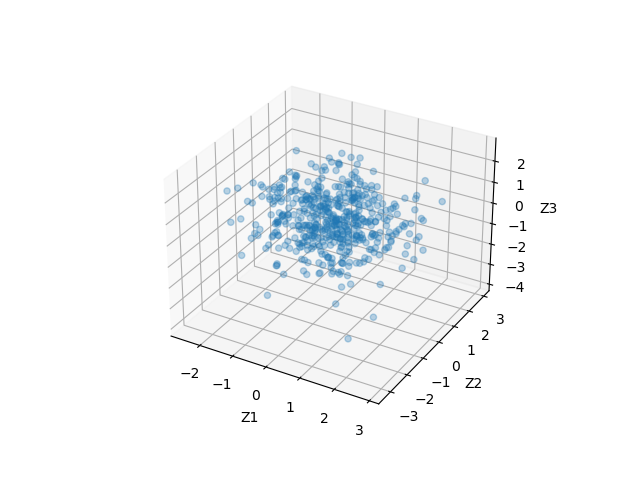}
    \end{subfigure}
    \begin{subfigure}{0.34\textwidth}
        \includegraphics[width=\textwidth]{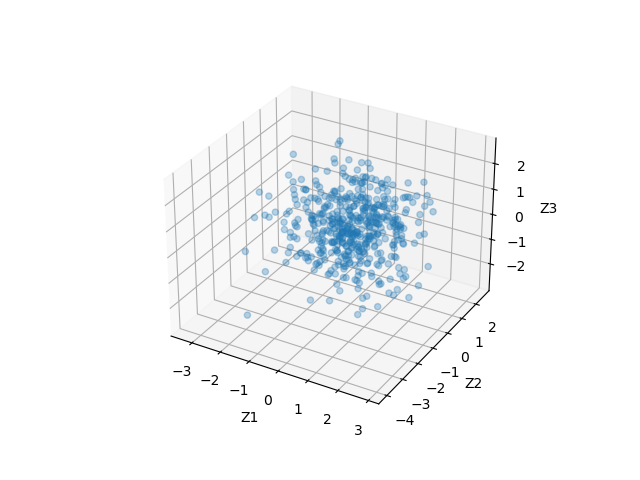}
    \end{subfigure}
    \caption{Scatterplots of representations learned by the VAE in the 3d toy problem.}
    \label{fig:scatter-representations}
\end{figure}

In Figure \ref{fig:scatter-representations}, we scatterplot the representations learned by the VAE in the 3d toy problem. Their distribution is meant to be marginally standard Gaussian, as oppose to the ``observable" data from Figure \ref{fig:data}.

\begin{figure}
    \centering
    \begin{subfigure}{0.4\textwidth}
        \includegraphics[width=\textwidth]{figures/example_graph.png}
    \end{subfigure}
    \begin{subfigure}{0.4\textwidth}
        \includegraphics[width=\textwidth]{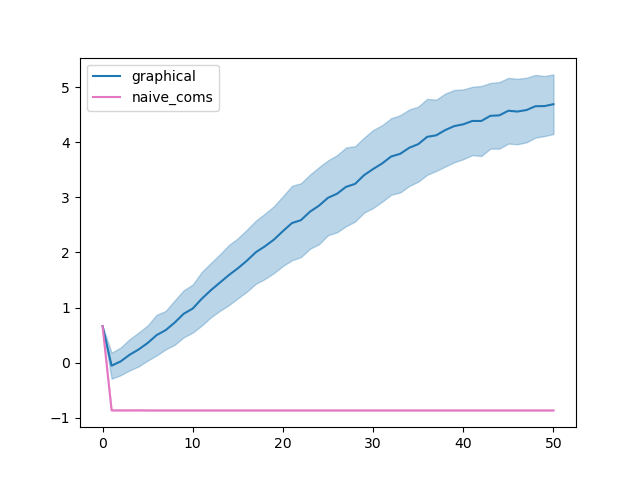}
    \end{subfigure}
    \caption{DIG of the ground-truth function in a 7-dimensional problem (left) and comparison Algorithm \ref{algorithm:gddo} (blue) vs. COMs (pink) in the task (right).}
    \label{fig:d7scores}
\end{figure}

In Figure \ref{fig:d7scores}, we compare Algorithm \ref{algorithm:gddo} to COMs on a 7-dimensional toy task.

\begin{figure}
    \centering
    \includegraphics[width=0.5\textwidth]{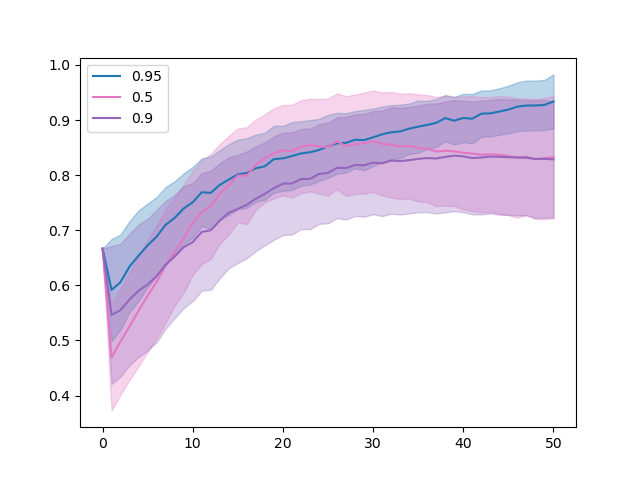}
    \caption{Performance of Algorithm \ref{algorithm:gddo} as the confidence level $1-\alpha$ varies.The fraction of invalid candidates produced by each algorithm}
    \label{fig:confidence}
\end{figure}

In Figure \ref{fig:confidence}, we study how the performance of Algorithm \ref{algorithm:gddo} depends on the confidence level $1-\alpha$ (the higher the sparser the recovered DIG), in SuperConductor. We can conclude that high confidence (sparsity) aids reliability of the learned model and improves optimization of the policy. In \ref{fig:invalid}, we plot the fraction of invalid candidates produced by each algorithm (Algorithm \ref{algorithm:gddo} and COMs) in the 3d toy task. That is, the proportion of such $\hat{\rvx}$ for which a ``groundt-truth" $\hat{\rvz}$ does not exist. While Algorithm \ref{algorithm:gddo} quickly learns the right manifold, COMs tends to produce invalid candidates.

\clearpage
\section{Hyperparameters For Experiments}
In all runs, for both Algorithm \ref{algorithm:gddo} and COMs \cite{trabucco2021conservative}, all neural networks were single-hidden layer MLPs, with 128 hidden units. Also, both methods were trained on batches of size 500. The rest of hyper-parameters: learning rates, gradient ascent steps, COMs gap (for COMs), for both COMs and our algorithm, were adopted from the COMs paper \cite{trabucco2021conservative}.

In the toy example, both methods were trained for $2e4$ steps. In SuperConductor, they were trained for $50$ epochs.

We estimate the running average of the pseudo-Hessian for Algorithm \ref{algorithm:ggd} with momentum parameter $0.99$. The edge test is done at $1-\alpha = 0.95$ confidence level.
In the toy example, we update the model's belief of the graph every $5e3$ steps, while in SuperConductor every $100$ steps.